\documentclass[twoside]{article}

\usepackage[accepted]{aistats2025}

\usepackage{amsmath}
\usepackage{amssymb}
\usepackage{bm}

\usepackage{graphicx}
\usepackage{placeins}
\usepackage{float}

\usepackage[ruled,vlined]{algorithm2e}

\usepackage{multirow}
\usepackage{enumitem}

\usepackage{xcolor}

\usepackage[round]{natbib}
\renewcommand{\bibname}{References}
\renewcommand{\bibsection}{\subsubsection*{\bibname}}

\usepackage{amsmath,amsfonts,bm}

\def\1{\bm{1}}

\def\rx{{\textnormal{x}}}

\def\rvw{{\mathbf{w}}}
\def\rvx{{\mathbf{x}}}
\def\rvy{{\mathbf{y}}}

\def\vc{{\bm{c}}}

\def\vw{{\bm{w}}}
\def\vx{{\bm{x}}}
\def\vy{{\bm{y}}}

\def\mA{{\mathbf{A}}}
\def\mB{{\mathbf{B}}}
\def\mC{{\mathbf{C}}}

\def\mG{{\mathbf{G}}}
\def\mH{{\mathbf{H}}}
\def\mI{{\mathbf{I}}}

\def\mL{{\mathbf{L}}}
\def\mM{{\mathbf{M}}}

\def\mS{{\mathbf{S}}}

\def\mU{{\mathbf{U}}}
\def\mV{{\mathbf{V}}}

\def\mY{{\mathbf{Y}}}

\newcommand{\T}{\mathsf{T}}

\newtheorem{lemma}{Lemma}
\newtheorem{theorem}{Theorem}
\newtheorem{corollary}{Corollary}
\newtheorem{definition}{Definition}
\newtheorem{proposition}{Proposition}

\begin{document}

%
\runningtitle{Learning Stochastic Nonlinear Dynamics with Embedded Latent Transfer Operators}

%
\runningauthor{Naichang Ke, Ryogo Tanaka and Yoshinobu Kawahara}

\twocolumn[

\aistatstitle{Learning Stochastic Nonlinear Dynamics with \\Embedded Latent Transfer Operators}

\aistatsauthor{Naichang Ke* \And Ryogo Tanaka* \And  Yoshinobu Kawahara}

\aistatsaddress{ The University of Osaka \\ naichang.ke@ist.osaka-u.ac.jp \And  The University of Osaka\\ryogo.tanaka@ist.osaka-u.ac.jp \And The University of Osaka \& RIKEN\\kawahara@ist.osaka-u.ac.jp} ]

\begin{abstract}\vspace*{-.8em}
We consider an operator-based latent Markov representation of a stochastic nonlinear dynamical system, where the stochastic evolution of the latent state embedded in a reproducing kernel Hilbert space is described with the corresponding transfer operator, and develop a spectral method to learn this representation based on the theory of stochastic realization. The embedding may be learned simultaneously using reproducing kernels, for example, constructed with feed-forward neural networks. We also address the generalization of sequential state-estimation (Kalman filtering) in stochastic nonlinear systems, and of operator-based eigen-mode decomposition of dynamics, for the representation. Several examples with synthetic and real-world data are shown to illustrate the empirical characteristics of our methods, and to investigate the performance of our model in sequential state-estimation and mode decomposition.\vspace*{-.7em}
\end{abstract}

\section{Introduction}
\label{sec:intro}

The problem of extracting the dominant dynamics and of predicting the behaviors of stochastic nonlinear dynamical systems from temporally sampled data is fundamental in a broad range of scientific and engineering fields that has been extensively investigated in the past, but still is technically challenging.

One of the most widely used techniques for forecasting in dynamical systems is the sequential Bayesian filter (a.k.a.\@ the Kalman filter), which was initially proposed by \citet{Kal60} and has been extended from various perspectives, in particular, to nonlinear systems, such as the extended Kalman filter~\citep{GA10}, the unscented Kalman filter~\citep{JU97,JU04}, the particle filter~\citep{Kit93,GSS93}, the kernel Kalman filter~\citep{GKN19}, and the one with neural networks~\citep{KSS17,KSBS17}.
Related to this, many methods have been discussed to learn nonlinear dynamical systems (state-space models) using sequential data, where one of the popular conventional approaches are the expectation-maximization algorithm \citep{GH96,GR19} and the variational approximations \citep{GH00} involving the estimation of latent state sequences. In machine learning, the so-called spectral method, a.k.a. subspace identification in control \citep{Kat05}, has been applied to this problem for several models, such as hidden Markov models \citep{HKZ09,SHSF09}. Both of these methods to learn nonlinear systems are, in principle, closely related; the latter can be regarded as an approximate estimate of the former (with robustness and lower computational cost).

In the meantime, operator-theoretic methods, such as Koopman analysis~\citep{Koo31,Mez13,BBKK22}, have recently attracted much attention in scientific fields due to the generality for various systems (such as Hamiltonian and dissipative systems), the connection to physical concepts~\citep{SKN17,MM18}, and advances of estimation methods such as dynamic mode decomposition (DMD)~\citep{RMB+09,Sch10,CTR12}. In particular, analysis of dynamical systems with Koopman operator (a.k.a.\@ composition operator) has received attention also in machine learning, and combined with machine learning techniques to develop more advanced methods for this problem~\citep{Kaw16,TKY17b,AELM20,KLNP23}. On the other hand, analyses with transfer operator (a.k.a.\@ Perron-Frobenius operator), which is the conjugate of Koopman operator, have been discussed for a few decades to study complex dynamic processes of distributions, such as molecular and quantum dynamics \citep{DFJ00,FP09,KNK+18}.

In this paper, we consider an operator-based latent Markov representation of a stochastic nonlinear dynamical system, where the stochastic evolution of the latent state embedded in a reproducing kernel Hilbert space (RKHS) is described with the corresponding transfer operator, and then develop a spectral method to learn this representation based on the theory of stochastic realization. The embedding may be learned adaptively with kernels, for example, constructed with feed-forward neural networks. The proposed model is regarded as a generalized state-space representation for nonlinear stochastic processes with transfer operators on embedded latent states. In addition, we address the generalization of sequential state-estimation (Kalman filtering) in stochastic nonlinear systems, and of operator-based eigen-mode decomposition of dynamics, for this representation. The former provides a computationally-feasible generalization of the existing non-linear filtering algorithms using evolution models estimated with the proposed spectral method. And the latter proposes a robust and genetic alternative to estimate Koopman mode decomposition \citep{BMM12,Mez05} (or DMD) to extract a set of transformations of coordinates, where dynamics are expressed as linear systems. Finally, we show several numerical results with synthetic and real-world data to illustrate the empirical characteristics and to investigate the performance of our method in sequential state-estimation and mode decomposition.

The remainder of this paper is organized as in the following: In Section~\ref{sec:background}, we briefly review the Koopman and transfer operators, and the Hilbert space embeddings of distributions. Then, in Section~\ref{sec:elto}, we define the embedded latent transfer operator (ELTO), which describe the evolution of latent states. In Section~\ref{sec:learning}, we develop the theory of stochastic realization for the representation with ELTOs, and then present a spectral method to learn this model. In addition, we develop the sequential-state estimation procedure in Section~\ref{sec:estimation}, and an estimation method for Koopman mode decomposition in Section~\ref{sec:mode_decomp}, respectively for our model. Finally, we conclude this paper in Section~\ref{sec:conclusion}. All proofs and the experimental details are included in the supplementary material.

\section{Background}
\label{sec:background}

\subsection{Koopman and Transfer Operators}
\label{ssec:operators}



Let $\{\rvx(t)\}_{t\in\mathbb{T}}$ ($\mathbb{T}$$\,=$$\,\{0,\pm 1,\pm 2,\ldots\}$) be a discrete-time stationary and ergodic Markov process in some vector space $\mathbb{X}\subset\mathbb{R}^p$, which is often called the state space. Then, {\em the transition density function}, which we denote by $p_{\mathrm{tr}}$, is defined by
\begin{equation*}
\Pr(\rvx(t+1)\in\mathbb{A}|\rvx(t)=\vx)=\int_\mathbb{A} p_\mathrm{tr}(\bm{z}|\vx)d\bm{z},
\end{equation*}
where $\mathbb{A}$ is any measurable set. The transition density function $p_\mathrm{tr}$ captures all information about the process $\rvx(t)$ by the stationary and Markov assumptions.

Now, we let $g\in\mathbb{G}(\mathbb{X})$ be an observable of the system, where $\mathbb{G}(\mathbb{X})$ is some function space over $\mathbb{X}$. Then, the corresponding {\em Koopman operator} (a.k.a.\@ {\em composition operator}) $\mathcal{K}\colon\mathbb{G}(\mathbb{X})\to\mathbb{G}(\mathbb{X})$ is defined by the following~\citep{Koo31,Mez05,KSM20}:
\begin{equation*}
\hspace*{-.5mm}\mathcal{K} g (\vx)
= \int p_\mathrm{tr}(\bm{z}|\vx) g(\bm{z}) d \bm{z}
= E[g(\rvx(t+1))|\rvx(t)=\vx].
\end{equation*}
That is, this expression gives the expectation of the observable at the next time step. Note that it is straightforward to see that the operator is linear (see, for example, the above references). Thus, we see that $\mathcal{K}$ acts linearly on the function $g$, even though the dynamics determined by $p_\mathrm{tr}$ may be nonlinear.

In the meantime, {\em the transfer operator} (a.k.a.\@ {\em Perron-Frobenius operator}) gives a representation of the system with respect to the density function of the state. Let $p_t\in \mathbb{L}^1(\mathbb{X})$ be a probability density function of the state vector at time $t$. Then, the transfer operator $\mathcal{T}\colon \mathbb{L}^1(\mathbb{X})\to \mathbb{L}^1(\mathbb{X})$ is defined by
\begin{equation*}
\mathcal{T} p_t(\vx) = \int p_\mathrm{tr}(\vx|\bm{z}) p_t(\bm{z}) d\bm{z}.
\end{equation*}
That is, the transfer operator pushes forward density $p_t$ to $p_{t+1}$, and consequently can be regarded as the linear counterpart of the transition probability in $\mathbb{L}^1(\mathbb{X})$.

Note that, in a measure-preserving and invertible system, the above both operators become unitary, and thus satisfies $\mathcal{K}=\mathcal{T}^{-1}$. This implies that the system can be equivalently described with both operators.

\subsection{Hilbert-Space Distribution Embeddings}
\label{ssec:embedding}

An embedding of the density over a random variable into RKHSs allows us to represent an arbitrary probability distribution non-parametrically by a potentially infinite dimensional feature vector, and to infer over those entirely in the space \citep{SGSS07,SHSF09}.
Let $\mathbb{M}_+^1(\mathbb{X})$ be the space of all probability measures $P$ on $\mathbb{X}$, and
$\rvx$ be a random variable taking values in $\mathbb{X}$ (an instantiation of $\rvx$ will be denoted by the corresponding roman character $\vx$). Also we let $k$ be a measurable positive definite kernel on $\mathbb{X}$ such that $\mathrm{sup}_{\vx\in\mathbb{X}}k(\vx,\vx')<\infty$, and $\mathbb{H}$ be the corresponding RKHS. Then, the kernel mean embedding $\mu_P\in\mathbb{H}$ for $P\in\mathbb{M}_+(\mathbb{X})$ exists, and is defined by a mapping $\mu\colon \mathbb{M}_+(\mathbb{X})\to\mathbb{H},\, P\mapsto \int \phi(\vx) dP(\vx)$, where $\phi$ is the corresponding feature map. In what follows, we sometimes use the notation $\mu(\rvx)$$\,:=\mu_{P_{\rvx}}$ for the embedding of the probability distribution $P_{\rvx}$ over a random variable $\rvx$. Different kernel functions result in different representations of the distribution on $\rvx$.
It is known that the kernel mean embedding $\mu_P$ fully characterizes the distribution if $k$ is a characteristic kernel \citep{FBJ04,SGF+08}.

In order to construct such an embedding, we may use a characteristic kernel function $k$, such as the Gaussian and Laplace kernel functions, or a learnable kernel endowed with some neural networks, for example, as in the same philosophy with the so-called deep \textcolor{black}{kernel} learning \citep{WHSX16}. That is, we employ a characteristic kernel function with inputs transformed by a neural network, i.e., $k_\theta(\vx,\vx') = k(\bm{f}_\theta(\vx),\bm{f}_\theta(\vx'))$, where $\bm{f}_\theta$ is a neural network with parameters $\theta$. The parameters $\theta$ in the transformation $\bm{f}_\theta$ are learned simultaneously with other parameters for a target task based on a criterion respectively designed for the task.


In addition, the concept of the covariance is also generalized based on the RKHS embeddings. First, we let $(\rvx, \rvy)$ be a random variable taking values in $\mathbb{X}\times\mathbb{Y}$ with corresponding marginal distribution $P_\rvx$ and $P_\rvy$, respectively, and joint distribution $P_{\rvx, \rvy}$. And, we let $\mathbb{H}$ and $\mathbb{G}$ be the respective RKHS associated with these variables for some appropriately-chosen kernels. Then, the covariance operator $\mathcal{C}_{\rvx}$ and cross-covariance operator $\mathcal{C}_{\rvy\rvx}$ are defined as follows \citep{baker1973}:
\begin{definition}
    Let $\phi$ and $\psi$ be feature maps associated with RKHSs $\mathbb{H}$ and $\mathbb{G}$, respectively. Then $\mathcal{C}_\rvx:\mathbb{H} \rightarrow \mathbb{H}$ and $\mathcal{C}_{\rvy\rvx}:\mathbb{H} \rightarrow \mathbb{G}$ are defined as
    \begin{align*}
        \mathcal{C}_\rvx &:=\int \phi(\bm{x})\otimes\phi(\bm{x})dP_\rvx(\bm{x}) \\
        \mathcal{C}_{\rvy\rvx} &:=\int \psi(\bm{y})\otimes\phi(\bm{x})dP_{\rvx, \rvy}(\bm{x}, \bm{y}). 
    \end{align*}
\end{definition}
Then, the cross-covariance of two functions $f\in\mathbb{H}$ and $g\in\mathcal{\rvy}$ can be formulated as
\[
    E_{\rvx\rvy}[f(\rvx)g(\rvy)] =\left<f, \mathcal{C}_{\rvx\rvy}g\right>_{\mathbb{H}} = \left<\mathcal{C}_{\rvy\rvx}f, g\right>_{\mathbb{G}}.
\]
From the above, the conditional mean embedding $\mu_{\rvy|\rvx}$, which is the embedding of a conditional distribution on a RKHS, is defined as follows \citep{SHSF09}:
\begin{definition}
    Let $\mathcal{C}_\rvx$ be the covariance operator on $\mathbb{H}$ and $\mathcal{C}_{\rvy\rvx}$ be he cross-covariance operator from $\mathbb{H}$ to $\mathbb{G}$. Assume $E_{\rvy|\rvx}[g(\rvy)|\rvx=\cdot]\in \mathbb{H}$ for all $g\in \mathbb{G}$, it follows for all $g\in\mathbb{G}$:
    \[
        E_{\rvy|\rvx}[g(\rvy)|\rvx=\bm{x}] = \left< g, \mathcal{C}_{\rvy\rvx}\mathcal{C}_\rvx^{-1}k(\bm{x}, \cdot ) \right>_{\mathbb{G}}.
    \]
    Then, the conditional mean embedding is defined by
    \[
    \mu_{\rvy|\rvx}:=\mathcal{C}_{\rvy\rvx}\mathcal{C}_\rvx^{-1}k(\bm{x}, \cdot ).
    \]
\end{definition}
Note that the corresponding operator $\mathcal{C}_{\rvy|\rvx}:=\mathcal{C}_{\rvy\rvx}\mathcal{C}_\rvx^{-1}$ is called as {\em the conditional covariance operator}.

\section{Embedded\,Latent\,Transfer\,Operator}
\label{sec:elto}

As we mentioned above, the transition probability, or the corresponding transfer operator, captures all information about a stochastic process if it is stationary and Markovian. However, a process of interest (denoted by $\rvy(t)$ here), where we obtain samples through some observation, may not satisfy this assumption in general. Therefore, here we consider the transfer operator for a (latent) process, which is obtained by mapping $\rvy(t)$ so that the mapped one satisfies the stationary and Markovian assumptions.

\paragraph{Stochastic Observation:}\hspace*{-3mm}
As is common in many existing works on the state estimation in dynamical systems, we consider {\em an observation probability distribution} $p_\mathrm{ob}$, which is defined for any measurable set $\mathbb{A}'$ by $\Pr(\rvy(t)\in\mathbb{A}'|\rvx(t)=\vx)=\int_{\mathbb{A}'}p_{\mathrm{ob}}(\vy|\vx)d \vy$. In the context of the operator representation above, this can be modeled by defining $g(\vx) = \int p_{\mathrm{ob}}(\vy|\vx)\vy d \vy$ for the observable $g$, where the corresponding Koopman operator is given by
\begin{equation*}
\begin{split}
\mathcal{K} g(\vx) &= \int \int p_\mathrm{tr}(\bm{z}|\vx) p_{\mathrm{ob}}(\vy|\bm{z}) \vy d \bm{z} d \vy \\
 &=  E[\rvy(t+1) | \rvx(t)=\vx].
\end{split}
\end{equation*}
Alternatively, this stochastic observation process can be described with an operator $\mathcal{O}\colon \mathbb{L}^1(\mathbb{X})\to \mathbb{L}^1(\mathbb{X})$, which we refer as {\em an observable operator}, defined by
\begin{equation*}
\mathcal{O} p(\vy) = \int p_{\mathrm{ob}}(\vy|\vx)p(\vx) d\vx.
\end{equation*}

\paragraph{Embedded Latent Markov Process:}
We suppose that there exists a latent process $\rvx(t)$ satisfying the stationary and ergodic Markovian assumptions, and that $\rvy(t)$ is obtained via some unknown stochastic observation from $\rvx(t)$. Then, we first embed the density of $\rvx(t)$ into the RKHS using the kernel mean embedding $\mu_{P_{\rvx(t)},\theta}$, for a characteristic kernel $k_{\theta}$ with parameters $\theta$, and then consider the transfer operator for the embedded process. Under the existence of such a latent process, the transition density function, or the corresponding transfer operator, could capture all information to recover the process $\rvy(t)$ if the observation mapping is well-estimated.
\begin{definition}[\rm Embedded Latent Transfer Operator]
Consider a process $\rvy(t)$ obtained through a stochastic observation from a (latent) process $\rvx(t)$. Then we call the transfer operator, w.r.t. the transition probability on $\rvx(t)$ and the embedded process $\mu_{P_{\rvx(t)},\theta}$, the Embedded Latent Transfer Operator (ELTO) for $\rvy(t)$ and denote it by $\mathcal{T}_{e,\theta}$.\vspace*{-.1em}
\end{definition}
That is, $\rvy(t)$ is regarded as a process obtained via some {\em unknown} observation from a (latent) stationary and ergodic Markov process $\rvx(t)$, which is fully characterized by the ELTO $\mathcal{T}_{e,\theta}$. Similarly, we consider the observable operator for the embedded densities of $\rvx(t)$ and $\rvy(t)$ respectively by the mean embeddings, which we call {\em an embedded observable operator (EOO)} and denote it by $\mathcal{O}_{e,\theta}$.
Then, from the definitions, the ELTO and EOO can be given by
\begin{equation}
\label{eq:elto_obs}
\mathcal{T}_{e}=\mathcal{C}_{\rvx(t+1)|\rvx(t)}
~~\text{and}~~
\mathcal{O}_{e} = \mathcal{C}_{\rvy(t)|\rvx(t)},
\end{equation}
where $\theta$ is abbreviated from the notations (hereafter, we abbreviate $\theta$ unless otherwise required).

\begin{figure}[t]
\centering
 \vspace*{-0em}
\includegraphics[width=.85\columnwidth]{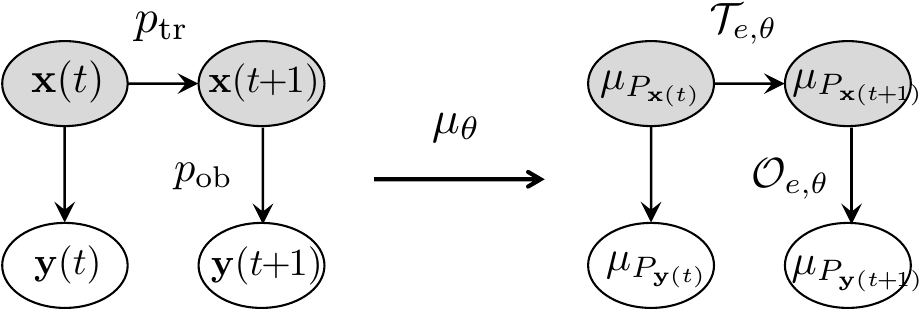}
\vspace*{-.2em}
\caption{Schematic overview of the embedded latent transfer operator (ELTO) $\mathcal{T}_{e,\theta}$.}\vspace*{-.9em}
\label{fig:elto}
\end{figure}

\section{Spectral Learning}
\label{sec:learning}

We consider the problem of estimating $\mathcal{T}_e$ (and $\mathcal{O}_e$) to capture the dynamics of a stochastic process of interest $\{\rvy(t)\}_{t\in\mathbb{T}}$ using data $\mY:=[\vy_0,\vy_1,\ldots,\vy_T]$, where $\vy_\tau:=\vy(t_0+\tau)$ $(\tau=0,1,2,\ldots,T)$ for some $t_0\in\mathbb{T}$ sampled from the process.\footnote{Here we assume that data is obtained as a single sequence that is consecutive in time for the simplicity. However, data used are not necessarily such a single sequence but a collection of sequences.} 
Note that, as mentioned above, $\rvy(t)$ does not necessarily need to be a Markov process, and we aim to gain a Markovian representation with a latent process to approximate it as in the same policy with approaches using state-space models. To this end, we develop a formulation of spectral learning, a.k.a.\@ subspace identification in system control, in analogy with the one based on stochastic realization.

\subsection{Stochastic Realization}
\label{ssec:stocha}

The theory of stochastic realization provides a method of constructing a Markovian representation for a stationary stochastic process with a prescribed covariance~\citep{Aka75,DPK85}. This theory has been applied to system identification in system control (known as {\em subspace identification}) \citep{LP96,KP99,Chi05}, and also to the problem of learning dynamical systems, such as hidden Markov models, in the machine learning field, where the approach is known as {\em spectral method} \citep{KYM07,HKZ09,SBS+10}. Here, we employ stochastic realization to obtain a Markovian representation with the ELTO $\mathcal{T}_{e}$ for the process $\rvy(t)$.

Consider a stationary process $\rvy(t)$ defined on $\mathbb{Y}$$\,\,\subseteq$$\,\,\mathbb{R}^q$. Let $k_y$ be a characteristic kernel on $\mathbb{Y}$, and denote the associated RKHS and feature map respectively by $\mathbb{H}_y$ and $\phi_y$. We assume that the process is with zero mean and covariance $\mathrm{Cov}(l)=E[u(\rvy(t$$\,+$$\,l))u(\rvy(t))]$, where $u\in\mathbb{H}_y$ and the covariance satisfies $\sum_{l=-\infty}^\infty\|\mathrm{Cov}(l)\|<\infty$.\footnote{Note that the covariance becomes a function of the time-lag $l$ because of the stationary assumption.} Note that $\mathrm{Cov}(0)>0$ holds due to $u\in\mathbb{H}_y$, which means $u(\rvy(t))$ is regular in the sense of \citep{HD88,Roz67}. Then the Hilbert space by $u(\rvy(t))$ is defined as follows. We first denote the space by all finite linear combinations of $u(\rvy(t))$ by\vspace*{-1.7mm}
\begin{center}
\hspace*{-1.5mm}$\mathbb{S}_y$$\,:=$$\,\left\{ {\textstyle \sum_{k=k_1}^{k_2}} c_k u(\rvy(k))|c_k\in\mathbb{R}\right\} (-\infty$$\,\leq$$\,k_1$$\,<$$\,k_2$$\,<$$\,\infty)$,\hspace*{-2mm}\vspace*{-2mm}
\end{center}
and then define an inner product between elements $\xi$$\,=$$\,\sum_{i=i_1}^{i_2}$$a_i u(\rvy(i))$ and $\zeta$$\,=$$\,\sum_{j=j_1}^{j_2}$$b_j u(\rvy(j))$ ($i_1\leq i_2$, $j_1\leq j_2$) by\vspace*{-2.3mm}
\begin{equation*}
\begin{split}
\left<\xi,\zeta\right>_{\mathbb{S}_y}
&=\mathrm{Cov}\left[{\textstyle \sum_{i=i_1}^{i_2}}a_i u(\rvy(i)), {\textstyle \sum_{j=j_1}^{j_2}}b_j u(\rvy(j))\right] \\
&={\textstyle \sum_{(i,j)\in \mathbb{I}}}a_ib_j \mathrm{Cov}(i-j),
\end{split}
\end{equation*}
where $\mathbb{I}=\{(i,j)\,|\, i_1\leq i\leq i_2, j_1\leq j\leq j_2\}$ is a finite set of indices. From the non-negative definiteness of the covariance function $\mathrm{Cov}(l)$, the above inner product defines a norm in $\mathbb{S}_y$, and thus $\mathbb{S}_y$ gives a Hilbert space by completing it with respect to the norm.

Now, we define the Hilbert subspaces generated by the {\em past} and {\em future} processes at time $t$ respectively by
\begin{equation*}
\begin{split}
\mathbb{S}_{t}^- &:= \overline{\mathrm{span}}\{u(\rvy(t-1)),u(\rvy(t-2)),\ldots\}
~~\text{and}~~\\
\mathbb{S}_{t}^+ &:= \overline{\mathrm{span}}\{u(\rvy(t)),u(\rvy(t+1)),\ldots\},
\end{split}
\end{equation*}
where $\overline{\mathrm{span}}$ means the closure of the space spanned by the linear combination of its elements. Then, an optimal predictor of $u(\rvy(t$$\,+$$\,j$$\,-$$\,1))$ ($j$$\,=$$\,1,\ldots,\infty$) based on the past $u(\rvy(t$$\,-$$\,i))$ ($i$$\,=$$\,1,\ldots,\infty$), in the sense of minimum variance, is obtained as the orthogonal projection of $u(\rvy(t$$\,+$$\,j$$\,-$$\,1))$ onto $\mathbb{S}_{t}^-$, which we denote by 
$\mathcal{P}(u(\rvy(t$$\,+$$\,j$$\,-$$\,1))|\mathbb{S}_{t}^-)$. In another word, the least-squares prediction of $u(\rvy(t$$\,+$$\,j$$\,-$$\,1))$ based on the past is given as an element of $\mathbb{X}_t:=\mathcal{P}(\mathbb{S}_{t}^+|{\mathbb{S}_{t}^-})$. The space $\mathbb{X}_t$ becomes finite dimensional under the assumption that the cross-covariance matrix $\mH$ of the future and past, i.e., $H_{i,j}$$:=\mathrm{Cov}(i$$-$$j$$+$$1)$ has finite rank. Moreover, if we call a subspace $\mathbb{D}$ ($\subset\mathbb{S}_{t}^-$) that satisfies $\mathcal{P}(\mathbb{S}_{t}^+|\mathbb{D})=\mathcal{P}(\mathbb{S}_{t}^+|\mathbb{S}_{t}^-)$ {\em a splitting subspace} \citep{LP91}, the following holds for $\mathbb{X}_t$:\vspace*{-2mm}
\begin{lemma}
\label{lem:finite}
Suppose that $\mathrm{rank}(\mH)<\infty$. Then, $\mathbb{X}_t$ is finite dimensional.\vspace*{-2mm}
\end{lemma}
\begin{lemma}
\label{lem:splitting}
$\mathbb{X}_t$ is a minimal splitting subspace for $\mathbb{S}_t^-$ and $\mathbb{S}_t^+$. In other words, for any splitting subspace  $\mathbb{D}$, it holds $\mathbb{D}\supset \mathbb{X}_t$.
\end{lemma}
\vspace*{-2mm}The proofs of these lemmas follow directly from the original ones in \citep{Kat05,LP91} because the difference only comes from the definitions of the covariances (thus, the calculation of orthonormal projections), which does not affects the proofs. 
These lemmas show that $\mathbb{X}_t$ contains the minimal necessary information to predict a future observation based on the past ones, and thus can be viewed as an interface between the past and future in stochastic processes. Therefore, a set of bases that spans $\mathbb{X}_t$ is available to construct a Markov model to optimally (in the least squares sense) approximates $u(\rvy(t))$ (and $\rvy(t)$), although its selection is not unique.

One of such a set of the bases is found by considering (kernel) canonical correlation analysis (CCA) between $\rvy_p(t)$$:=$$\{\rvy(t$$-$$i)\,|\,i$$=$$1,2,\ldots\}$ and $\rvy_f(t)$$:=$$\{\rvy(t$$+$$j$$-$$1)\,|$ $j$$=$ $1,2,\ldots\}$, as described in the following proposition (refer, for example, \citep{EH08} for CCA between processes):
\begin{proposition}
\label{prop:markov}
Assume that $\mathrm{rank}(\mH)$\,$=$\,$r$ ($<$\,$\infty$). Let $\sum_{i}\alpha_{l,i}u(\rvy(t-i))$ ($i=1,2,\ldots,$) be the $l$-th canonical direction for $u(\rvy_p(t))$ w.r.t.\@ the CCA between processes $u(\rvy_p(t))$ and $u(\rvy_f(t))$, where $\alpha_{l,i}\in\mathbb{R}$. Then,
\begin{equation*}
\mathbb{X}_t = \overline{\mathrm{span}}\{s_l^{1/2}{\textstyle \sum_i}\alpha_{l,i}u(\rvy(t-i))\,|\,l=1,\ldots,r\},
\end{equation*}
where $s_l$ is the $l$-th eigenvalue.
\end{proposition}

If we define $\rvx(t):=[\rx_1(t),\ldots,\rx_r(t)]^{\T}$, where $\rx_l (t):=$ $s_l^{1/2}\sum_i\alpha_{l,i}u(\rvy(t-i))$, then we have the following:
\begin{theorem}
\label{thm:realization}
Let denote by $\mathcal{P}(\xi_f|\mathbb{S}_t^-)=\vc^\T \mathcal{Q}\rvx(t)$ ($\xi_f$$\in$ $\mathbb{S}_t^+$ and $\vc$ is the corresponding coefficients) the orthonormal projection of $\xi_f$ onto $\mathbb{S}_t^-$. And define $\mA:=\mathcal{Q}^\dagger\mathcal{Q}^\uparrow$, 
 where $\bullet^\uparrow$ is the operation that removes the first block row. Then we have
\begin{equation*}
\rvx(t+1) = \mA \rvx(t) + \rvw(t),
\end{equation*}
where $\rvw(t):=\rvx(t+1)-\mathcal{P}(\rvx(t+1)|\rvx(t))$ ($\perp\mathbb{S}_t^-$).
\end{theorem}
Note that the above representation gives the evolution of a set of axes of $\mathbb{X}_t$ along time, which implies that a sequence of elements of $\mathbb{X}_t$ along time (a process) can be Markovian.

\subsection{Empirical Estimation}
\label{ssec:empirical}

We formulate the estimation of $\mathcal{T}_e$ using finite data $\mY$ based on the above results. For some integer $h\in(0,(T+1)/2)$, we first define\vspace*{-.2mm}
\begin{equation*}
\begin{split}
\bm{y}_{p,n} &:= [\bm{y}_{n+h-2}^\T,\ldots,\bm{y}_{n-1}^\T]^\T \\
\bm{y}_{f,n} &:= [\bm{y}_{h+n-1}^\T,\ldots,\bm{y}_{2h-2+n}^\T]^\T
\end{split}
\end{equation*}
for $n\,$$=$$\,1,\ldots,N$, where $N$$\,:=T$$\,+$$\,2$$\,-$$\,2h$. These are regarded as samples of $\rvy_p(t)$ and $\rvy_f(t)$, respectively. Also, the empirical embeddings of the densities of $\rvy_p(t)$ and $\rvy_f(t)$ are obtained as\vspace*{-.2mm}
\begin{equation*}
\begin{split}
    \widehat{\mu}(\rvy(t-i))&={\textstyle\sum_{n=1}^N} \phi_y(\bm{y}_{n+\tilde{i}}), \\
    \widehat{\mu}(\rvy(t+j-1))&={\textstyle\sum_{n=1}^N} \phi_y(\bm{y}_{n+\tilde{j}})
\end{split}
\end{equation*}
for $i,j$$\,=\,$$1,$$\,\ldots,$$\,h$, where $\tilde{i}\,$$=\,$$h$$\,-$$\,i-1$ and $\tilde{j}\,$$=\,$$h$$\,+$$\,j$$\,-$$\,2$.

Let $\mC_{pp}$ and $\mC_{ff}$ be the empirical covariance matrices of $u(\rvy_p(t))$ and $u(\rvy_f(t))$, respectively, and $\mC_{fp}$ the empirical cross-covariance matrix between those, which are calculated through the definition of the covariance $\mathrm{Cov}(l)=E[u(\rvy(t+l))u(\rvy(t))]$ (cf.~Appendix~B.1).
And we denote the square roots of $\mC_{ff}$ and $\mC_{pp}$ by $\mC_{ff}$$=$$\mL \mL^\T$ and $\mC_{pp}$$=$$\mM \mM^\T$, respectively. Then, $\bm{\alpha}_l$ (in Proposition~\ref{prop:markov}) can be calculated via the (truncated) singular-value decomposition (SVD) of $\mL^{-1}\mC_{fp}(\mM^{-1})^\T\approx \widehat{\mU}\widehat{\mS}\widehat{\mV}^\T$ as the $l$-th row of 
the matrix $\widehat{\mV}^\T \mM^{-1}$. Therefore, the $l$-th basis of $\mathbb{X}_t$ is estimated as $\sum_{i=1}^h b_{l,i}u(\rvy(t$$-$$i))$, where $b_{l,i}$ is the $(l,i)$-element of $\mB:=\widehat{\mS}^{1/2}\widehat{\mV}^\T \mM^{-1}$ and $u$ can be taken as
\begin{equation*}
u={\textstyle \sum_{e_i\in S}}w_i \phi_y(\bm{y}_{e_i}) \, (=:\Phi_S\bm{w})
\end{equation*}
for $\bm{w}\in\mathbb{R}^{|S|}$ ($\{e_1,\ldots,e_{|S|}\}=S\subseteq \{0,1,\ldots,T\}$) ($\Phi_S$ is the feature matrix). As a result, samples of $\rvx(t)$ that corresponds to $\bm{y}_{p,n}$ (and $\bm{y}_{f,n}$) are obtained as
\begin{equation*}
\bm{x}_n = \mB \Phi_{p,n}^\T \Phi_S \vw,
\end{equation*}
where $\Phi_{p,n}=[\phi_y(\bm{y}_{n+h-2}),\ldots,\phi_y(\bm{y}_{n-1})]$. Therefore, if we prepare a kernel $k_x$ on $\mathbb{R}^r$ for $\bm{x}_n$ and define the feature matrix $\Psi := [\phi_x(\bm{x}_1),\ldots,\phi_x(\bm{x}_N)]$, 
where $\phi_x$ is the corresponding feature map, then the ELTO and EOO are estimated as follows through the definitions~(\ref{eq:elto_obs}). That is, if we approximate the inverses of covariance operators with those regularized ones (cf.\@ \citep{FSG13}), those are given by
\begin{align*}
\widehat{\mathcal{T}}_e &= \Psi_2 \Psi_1^\T (\Psi_1 \Psi_1^\T + \epsilon\mathcal{I})^{-1} = \Psi_2 (\mG_1 + \epsilon \mI_{N-1})^{-1}\Psi_1^\T, \\
\widehat{\mathcal{O}}_e &= \Phi \Psi^\T (\Psi \Psi^\T + \epsilon\mathcal{I})^{-1} = \Phi (\mG_{yx}+\epsilon\mI_N)^{-1}\Psi^\T,
\end{align*}
where $\Psi_1 :=\Psi_{:,1:N-1}$, $\Psi_2:=\Psi_{:,2:N}$, $\Phi:=[\phi_y(\bm{y}_h),\ldots,$ $\phi_y(\bm{y}_{h+N-1})]$, $\mG_1=\Psi_1^\T\Psi_1$, $\mG_{yx}=\Phi^\T\Psi$, $\epsilon>0$, and $\mathcal{I}$ and $\mI$ are the identity operator and matrix.

The pseudo-code for the above spectral learning is found in Appendix~B in the supplementary document.

\section{Sequential State Estimation}
\label{sec:estimation}

Sequential state-estimation, or sequential Bayesian filtering (Kalman filtering), is a popular method to calculate the posterior of the state
$p(\rvx(t)|\vy(\bar{t}_0),\ldots,\vy(t))$ ($\bar{t}_0<t$) in a dynamical system using sequentially sampled data. This method is performed by sequentially calculating two updates of the state estimates: the {\em prediction} and {\em innovation} updates. In the prediction update, we propagate the {\em belief} state in time by applying the transition model. And, on new observations, the innovation update applied the Bayes' theorem to the {\em a-priori} belief state to obtain the {\em a-posteriori} one.

Now, we describe the prediction and innovation updates with the ELTO and EOO, whose derivations essentially follow those of the kernel Kalman rule (KKR) \citep{GKN19}. Hereafter, we denote by $\bullet^-$ and $\bullet^+$ the a-priori and posteriori beliefs (mean embeddings and covariances), i.e., the ones before and after getting new observation $\bm{y}(t)$, respectively. First, in the prediction update, the embedded state and its corresponding covariance operator are propagated in time by applying the ELTO, i.e.,
\begin{equation*}
\mu(\rvx(t))^- = \mathcal{T}_e\mu(\rvx(t-1))^+,
\mathcal{C}_{\rvx(t)}^- = \mathcal{T}_e\mathcal{C}_{\rvx(t-1)}^+\mathcal{T}_e^\T+\mathcal{C}_V,
\end{equation*}
where $\mathcal{C}_V$ is the covariance of the noise in the state evolution. Then in the innovation update, these are further updated by incorporating information of new observation $\bm{y}(t)$. For the embedded state, we have
\begin{equation*}
\mu(\rvx(t))^+ = \mu(\rvx(t))^- + \mathcal{G}_t \left(\phi(\bm{y}(t))-\mathcal{O}_e\mu(\rvx(t))^-\right),
\end{equation*}
where $\mathcal{G}_t\colon\mathbb{H}\to\mathbb{H}$ is the operator corresponding to the Kalman gain and we call it {\em the Kalman gain operator}. This is determined by minimizing the covariance of the error in the posterior estimation of observations, and given by
\begin{equation*}
\mathcal{G}_t = \mathcal{C}_{\rvx(t)}^- \mathcal{O}_e^\T \left(\mathcal{O}_e\mathcal{C}_{\rvx(t)}^-\mathcal{O}_e^\T+\mathcal{C}_W\right)^{-1},
\end{equation*}
where $\mathcal{C}_W$ is the covariance of the noise in the observation process. Then, the posteriori covariance operator for the state is updated as
\begin{equation*}
\mathcal{C}_{\rvx(t)}^+ = \mathcal{C}_{\rvx(t)}^- - \mathcal{G}_t\mathcal{O}_e\mathcal{C}_{\rvx(t)}^-.
\end{equation*}


Since the ELTO and EOO are unknown beforehand in general, those in the above updates need to be replaced with the empirical ones estimated by the procedures in Section~\ref{sec:learning}. Note that \citep{GKN19} provides no explicit procedure for estimating these operators.



\subsection*{Numerical Results}
We investigated the performance of our method for sequential Bayesian estimation in the following three cases:\@ Simulated single pendulum system, human motion dynamics (HuMoD) and image sequences of quad-link pendulum, under comparison with some benchmark methods. The experimental details are described in Appendix~C (in the supplementary document), and all the codes for producing the results shown here are included in the supplemental materials.

\vspace*{-1mm}\paragraph{Pendulum:}
Out first experiment is performed using a simulated pendulum system. We simulated a single pendulum with randomly initialized angles in the range $[-0.25\pi, 0.25\pi]$ and angular velocities in the range $[-2\pi/s, 2\pi/s]$.\footnote{The simulation was performed using the code in\\\texttt{https://github.com/gregorgebhardt/pyKKR}.} The pendulum dynamics were simulated with a frequency of 10,000 Hz, incorporating normally distributed process noise with standard deviation $\sigma = 0.1$. Observations of the joint positions were made with sampling rate of 10 Hz, with additive Gaussian noise $\mathcal{N}(q_t, 0.01)$ ($q_t$ is the angle at timestep $t$). In the experiment, we used the RBF Gaussian kernel for both $k_x$ and $k_y$ with the gamma's of $(1/\#\text{features})$. We also used the ADAM optimizer with a learning rate of 1e-3, with 200 training epochs for the latent state $x$, and the window size is set to 5. We further optimized the transition and regularization hyper-parameters using the Covariance Matrix Adaptation Evolution Strategy (CMA-ES) \citep{hansen2006cma}, with the initial step size of 0.5 for the search step magnitude in the parameter space.

\begin{figure}[t]
\centering
\vspace*{-1em}
\includegraphics[width=.9\columnwidth]{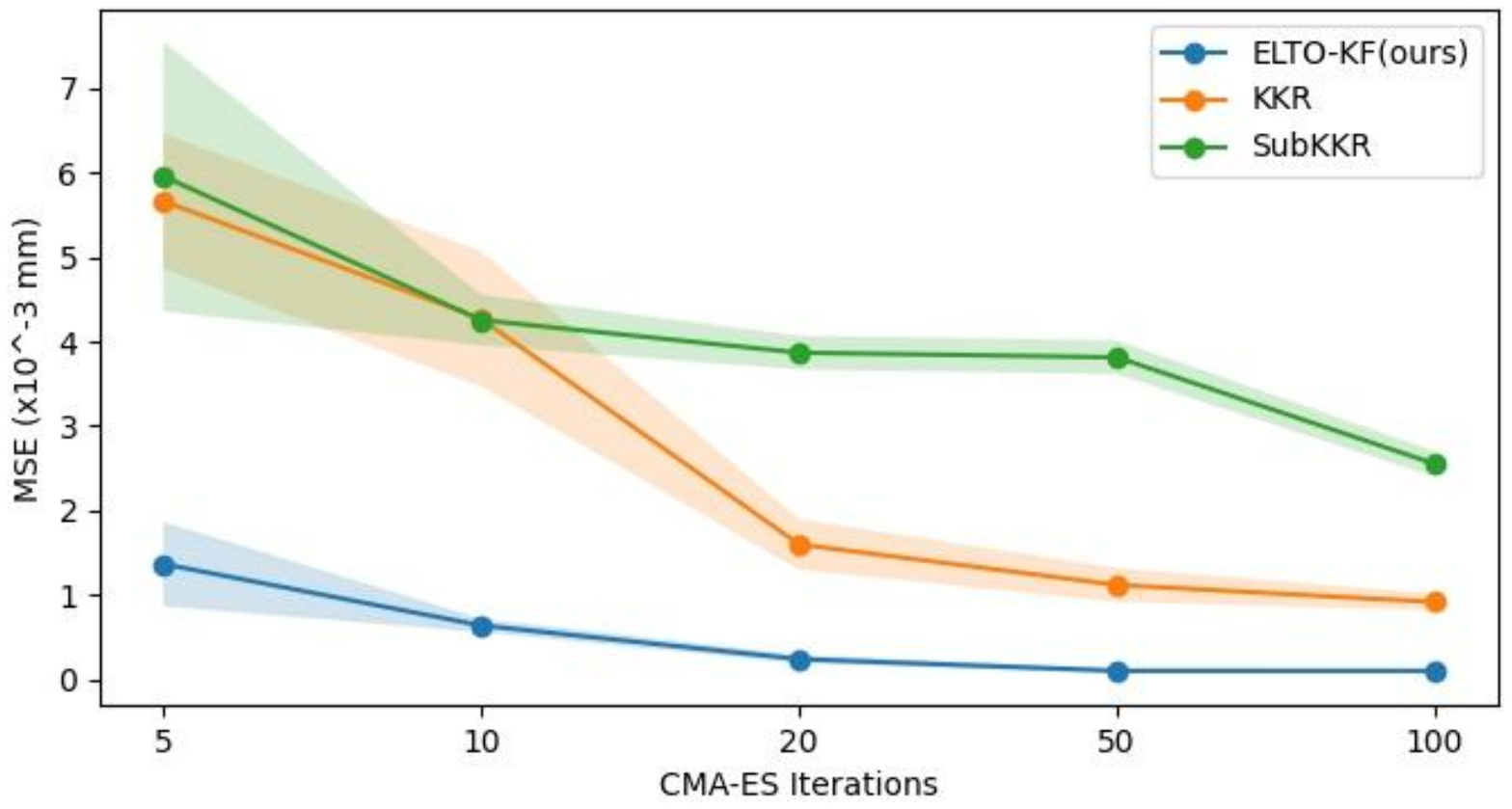}
\vspace*{-2.5em} 
\caption{Comparison of prediction performance by ELTO (ours) and KKR for the pendulum data.}
\label{fig:pendulum}
\end{figure}

Figure~\ref{fig:pendulum} depicts the mean-squared errors (MSEs) (one-step prediction) along the CMA-ES iterations by our method (ELTO) under the comparison with the kernel Kalman rule (KKR) and Subspace KKR (SUBKKR) \citep{GKN19} as benchmark methods. It seems that our method consistently outperforms the benchmarking method, demonstrating the capacity of our method to estimate more informative latent representation via the spectral learning, without requiring explicit provision of preceding states and observations (that are necessary for KKR and SUBKKR).

Additionally, we conducted the experiment under several different levels of process and observation noises to demonstrates the robustness of our model. The result is shown in Table~\ref{tab:noise-pen}, where the MSEs over five independent trials are depicted.

\begin{table}[t]
\centering
\caption{Model performance under different noise settings. $n_p$ and $n_o$ are the standard deviations of the process and observation noises, respectively.}
\label{tab:noise-pen}
\vspace*{.5em}
\begin{tabular}{cccc}
\hline
\multirow{2}{*}{Noise setting} & \multicolumn{3}{c}{MSE($\times 10^{-3}$)
} \\ \cline{2-4}
& SubKKR & KKR & ELTO(ours) \\ \hline
$n_p$=0.1, $n_o$=0.01 & 2.7574 & 1.1185 & \textbf{0.1010} \\ 
$n_p$=0.2, $n_o$=0.01 & 3.9321 & 3.4061 & \textbf{0.2325} \\ 
$n_p$=0.1, $n_o$=0.1 & 13.192 & 15.797 & \textbf{11.278} \\ 
\hline
\end{tabular}
\end{table}

\vspace*{-1mm}\paragraph{Human Motion Dynamics (HuMoD):}
The Human Motion Daynamics (HuMoD) dataset includes three dimensional coordinates of specific body parts captured by 36 markers, as well as electromyographic data from 14 sensors, covering a variety of motions performed by two subjects, such as steady walking and variable-speed running, and other dynamic activities \citep{wojtusch2015humod}. To assess the predictive capabilities of our method, we focused on the markers at the T8/T12 (8th and 12th thoracic vertebrae), leveraging the vertiacal position data (markerY in the HuMoD dataset) across different motion conditions. The original data, recorded at 500 Hz, was preprocessed by removing the initial idle period to exclude non-experimental influences, subsequently downsampled to 50 Hz, and then centralized to ensure uniformity. We used data from Subject B for the tranining and Subject A for the test, following the various motion tasks outlined in HuMoD. Also in the experiment, we used the RBF Gaussian kernel for both $k_x$ and $k_y$ with the gamma's of $(1/\#\text{features})$.

\begin{figure}[t]
\centering
\vspace*{-1em}
\includegraphics[width=.9\columnwidth]{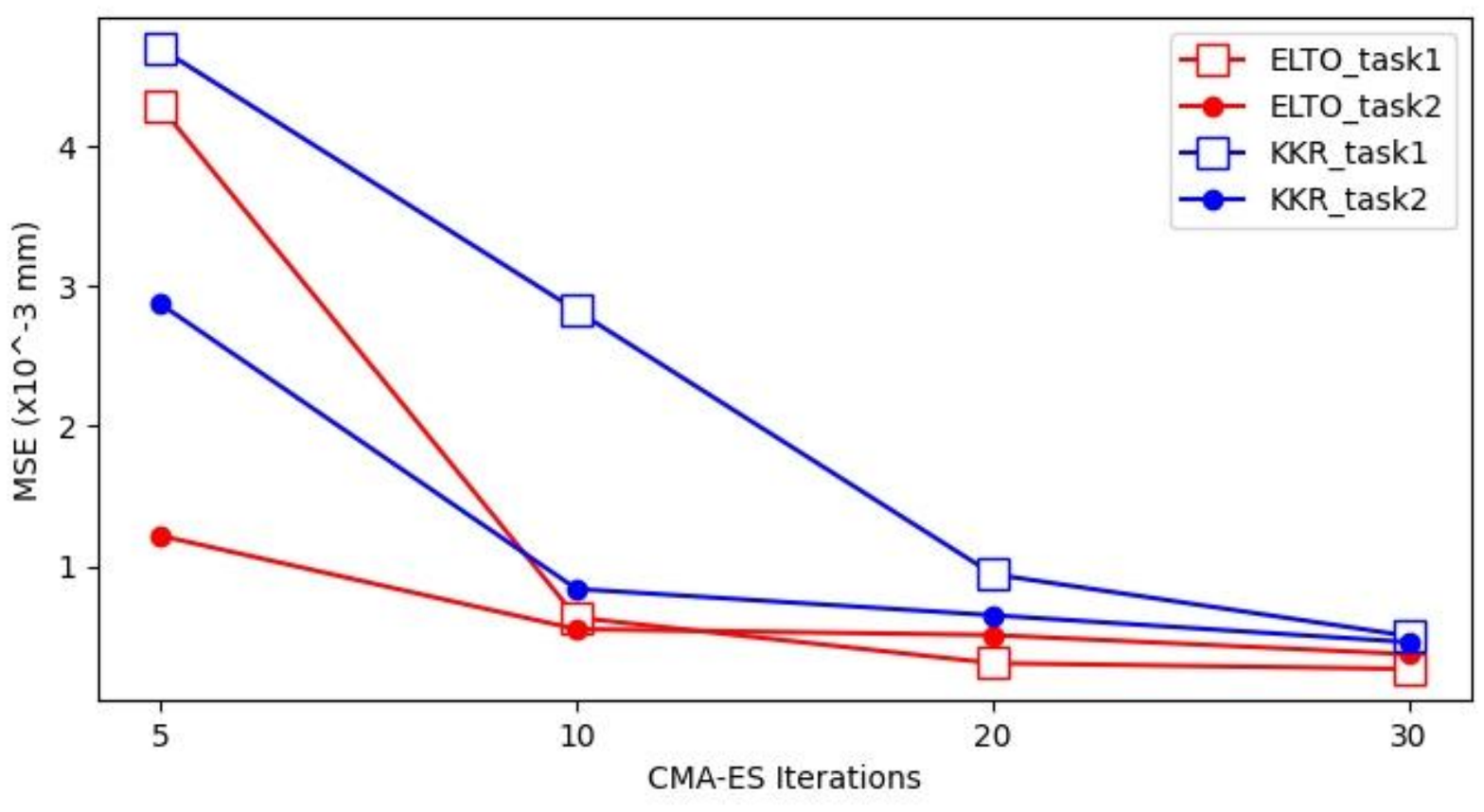}
\vspace*{-2.5em}
\caption{Comparison of prediction performance by ELTO (ours) and KKR for the HuMoD data. Task~1: Prediction of height of T8 when walking at 1.0 m/s and Task~2: Prediction of height of T12 when walking at 1.5 m/s.}
\label{fig:humod}
\end{figure}

Figure~\ref{fig:humod} depicts the MSEs (one-step prediction) along the CMA-ES iterations for two tasks: the prediction of height of T8 when walking at 1.0 m/s and of height of T12 when walking at 1.5 m/s, by our method (ELTO), under the comparison with KKR \citep{GKN19} as a benchmark method. As with in the previous experiment, it seems that our method consistently outperforms the benchmarking method, demonstrating the capacity of our method to estimate more informative latent representation via the spectral learning.

\vspace*{-3mm}
\paragraph{Image Sequence from a Quad-link Pendulum Simulation:}
Furthermore, we evaluated the capacity of our model using a simulated quad-link pendulum system, where the input data were in the form of $48 \times 48$ grayscale images.\footnote{The simulation was performed using the code in\\\texttt{https://github.com/LCAS/RKN/tree/master/n\_link\_sim}.} 
The quad-link pendulum system starts with initial positions sampled from $[-\pi, \pi]$, and all velocities are initialized to zero. Friction is applied to each link with a value of 0.1. The system evolves with a timestep of $10^{-4}$ seconds, and data is collected every 0.05 seconds. Optional noises are also added to the images, including occlusion and distortion. The noise parameters are set as $r=0.2$ with thresholds for noise application of $[0, 0.25, 0.75, 1.0]$ while the first 5 images in each sequence remain noise-free.
The dataset consists of different sequence lengths for training and testing.
In this experiment, we used a learnable kernel with a feedforward neural network $k_\theta(\vx,\vx') = k(\bm{f}_\theta(\vx),\bm{f}_\theta(\vx'))$ whose parameters $\theta$ were learned simultaneously in the spectral learning procedure. As for the encoder, we used a neural network with two Conv-ReLU-maxpooling and single fully connected layers. And for the decoders, a network with two transposed Conv-ReLU layers both for the mean and covariance. The details architectures are found in Appendix~C in the supplementary document. In the experiment, we first trained the learnable kernel in our model (ELTO-KF), followed by retraining the decoder to compare the output with the baseline method. Both training stages used the ADAM optimizer with a learning rate of 1e-3, where Gaussian non-negative likelihood was used for the training loss.

\begin{table}[t]
\centering
\vspace{-.5em}
\caption{Comparison of predictive performance on the experiment with quad-link pendulum images.}
\label{ta:quad}
\vspace{.5em}
{\small \begin{tabular}{lcc}
\hline
Model &  without noise & with noise \\ 
(train/test length) & MSE & MSE \\ \hline\hline
ELTO-KF (1.5k) & \textbf{0.2175} & \textbf{0.2942}  \\
RKN (1.5k)     & 0.2198 & 0.2944 \\
LSTM(1.5k)     & 0.2527 & 0.3167 \\\hline
ELTO-KF (15k)  & 0.2924 & \textbf{0.2936} \\
RKN (15k)      & \textbf{0.2838} & 0.3258  \\
LSTM(15k)      & 0.3050 & 0.3158 \\
\hline
\end{tabular}}
\end{table}

Table~\ref{ta:quad} shows the MSE by our model (ELTO-KF) under the comparison with the recurrent Kalman network (RKN) \citep{BPG+19} and LSTM as benchmark methods, which might not be the state-of-the-art methods but popular ones specialized for the prediction task, for datasets with several different setups. We could see that our model demonstrates comparable performance to the benchmark models, showing the capability of our method for forecasting although that is not necessarily a model specialized for this task.

\section{Mode Decomposition}
\label{sec:mode_decomp}

Operator-theoretic analysis, such as Koopman analysis, of dynamical systems has recently attracted much attention in various fields of scientific and engineering, as mentioned at the beginning of this paper. One of its major reasons is that linear algebraic calculations, such as spectral decomposition (eigen-decomposition), becomes available for the analysis of nonlinear systems due to the linearity of operator-based representations \citep{BMM12,BBKK22}. For example, using the eigen-decomposition of Koopman operator: $\mathcal{K}\varphi_j(\cdot)=\lambda_j\varphi_j(\cdot)$, where $\lambda_j$ and $\varphi_j$ are the $j$-th eigenvalue and eigenfunction, the evolution of an observable can be decomposed as in the following way:
\begin{equation}
\label{eq:kmd}
\mathcal{K}g(\cdot) = \sum_{j=1}^\infty \lambda_j \varphi(\cdot)v_j,
\end{equation}
where $v_j\in\mathbb{C}$. This is called {\em Koopman mode decomposition} (KMD), where each component in the summation has an inherent frequency and decay-rate determined by the eigenvalue. Note that dynamic mode decomposition (DMD) is regarded as a finite approximation of this decomposition using data.

The proposed spectral learning is useful to provide a robust way to calculate the KMD to capture the intrinsic dynamics in observed data. First, we consider the Koopman operator in the case where the observables mapped forward by that are elements of $\mathbb{H}$, and denote it by $\mathcal{K}_k$ (this is the one called {\em kernel Koopman operator} in \citep{KSM20}). Although this might not be satisfied in general, we introduce this in the expectation that this approximately well captures the {\em true} dynamics. Then, $\mathcal{K}_k$ can be given as follows, where $\mathcal{C}_{\rvx(t)}$ and $\mathcal{C}_{\rvx(t)\rvx(t+1)}$ are respectively the covariance operator for $\rvx(t)$ and cross-covariance operators $\rvx(t)$ and $\rvx(t+1)$.\vspace*{-2mm}
\begin{corollary}
If $g(\bm{x})=\int p_{\mathrm{ob}}(\bm{y}|\bm{x})\bm{y}d\bm{y}\in\mathbb{H}$, then we have $\mathcal{K}_k = \mathcal{C}_{\rvx(t)}^{-1}\mathcal{C}_{\rvx(t)\rvx(t+1)}$.\vspace*{-1mm}
\end{corollary}
The proof is found in Appendix~A in the supplementary document. Therefore, the KMD~(\ref{eq:kmd}) can be calculated via the eigen-decomposition of the finite matrix (refer, e.g.~\citep{MSKS20}) consisting of estimated state vectors by our spectral learning. 


\subsection*{Numerical Results}
In order to demonstrate the performance of our method for the mode decomposition, in particular, the robustness in estimating Koopman eigenvalues under process and observation noises, we applied our method to data generated from two popular nonlinear dynamical systems: Van der Pol (VDP) oscillator and Stuart-Landau (SL) oscillator, under comparison with the related conventional methods. The former is given by the following nonlinear system:
\begin{equation*}
    \frac{dx}{dt} = y, 
    \frac{dy}{dt} = \mu (1 - x^2) y - x,
\end{equation*}
where $\mu$ is the parameter controlling the nonlinearity, and the latter by
\begin{equation*}
    \frac{\partial z}{\partial t} = (\mu + i \gamma) z - (1 + i \beta) |z|^2 z + \sqrt{\epsilon} \eta
\end{equation*}
in the complex form, where $\mu$, $\gamma$, $\beta$ are the parameters and $\eta$ means system noise with covariance $\epsilon$. As comparison methods, we selected two popular ones: the Hankel DMD (hDMD) \citep{AM17} or extended DMD (eDMD) \citep{WKR15} and the subspace DMD (sDMD) \citep{TKY17a}. Note that sDMD is known to be a method with the capability of estimating a correct KMD under process noises with partial observation. In this experiment, we used the RBF Gaussian kernel for both $k_x$ and $k_y$ with 0.01 for our method (ELTO) after a grid search. And for $\Phi_S$ to estimate the state by our spectral learning, the last 600 samples were used.

\begin{figure}[t]
\centering
\vspace{-1em} 
\includegraphics[width=\columnwidth]{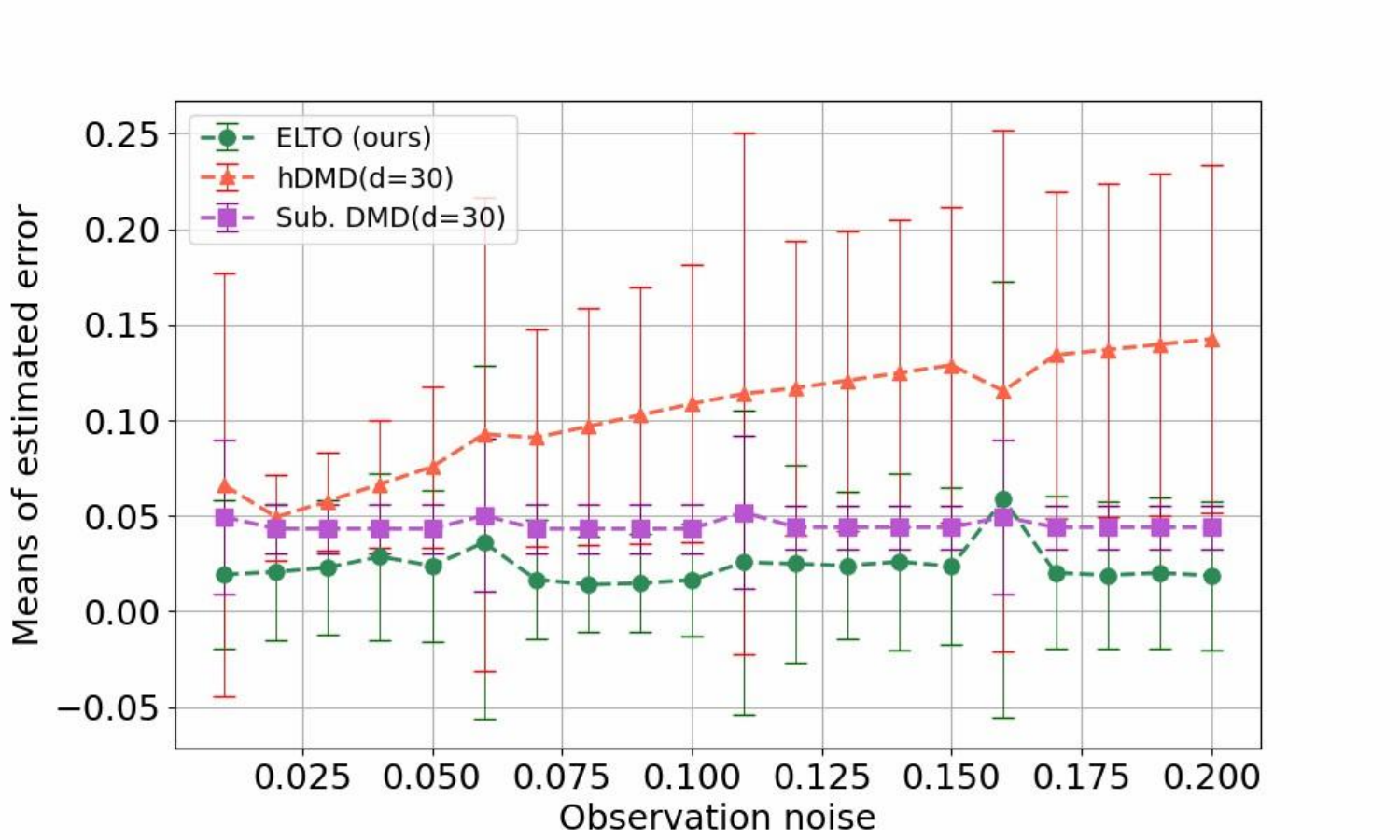}
\vspace{-2em} 
\caption{Results for VDP oscillator}
\label{fig:results_vdp}
\vspace{-.15em}
\includegraphics[width=\columnwidth]{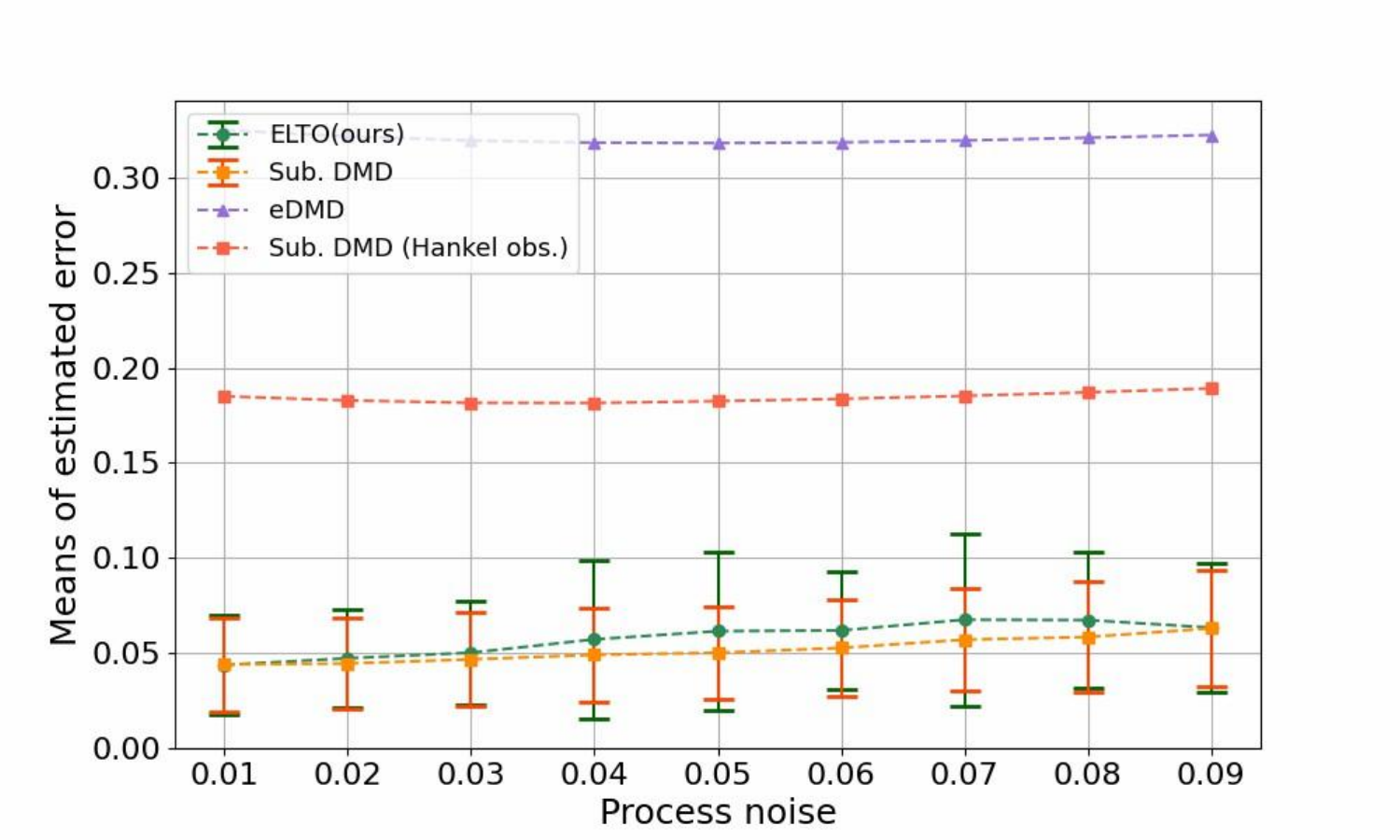}
\vspace{-2em} 
\caption{Results for SL oscillator}
\label{fig:results_SL}
\vspace{-1em}
\end{figure}

First, Figure~\ref{fig:results_vdp} depicts the estimation error of eigenvalues for the VDP oscillator by each method when the standard deviation of {\em observation noise} was varied from 0.01 to 0.2. As a comparison, we showed the results by hDMD  and sDMD with delay observations of 30 length. To evaluate the estimation error, the absolute error between the true eigenvalues and estimated ones was computed along the different observation noise levels. For the true eigenvalues, we numerically computed the frequency of the limit cycle from the simulation, adopted it as the fundamental frequency, and used integer multiples of the computed period as harmonics, utilizing the corresponding discrete eigenvalues at $dt = 0.1$. In the figure, the mean and standard deviation over 50 trials were calculated. We could see that our method (ELTO), similar to sDMD, shows smaller errors and greater robustness against observation noise intensity compared to deterministic DMD. Furthermore, ELTO demonstrates smaller average estimation errors than sDMD in most cases.

Next, Figure~\ref{fig:results_SL} depicts the estimation error for eigenvalues for the SL oscillator by each method when the standard deviation of {\em process noise} was varied from 0.01 to 0.09, with the observation noise having a variance of 0.01. As a comparison, we showed the results by sDMD with delay observations of 30 length, and eDMD, sDMD with arbitrary extension of observations. For the latter two comparison methods, we used observation via the polar representation of the form:
$\bm{y}_t = \left[ e^{-10i\theta_t}, e^{-9i\theta_t}, \dots, e^{9i\theta_t}, e^{10i\theta_t} \right]$,
which is used in \citep{TKY17a}. It should be noted that the system's eigenvalues, unlike in the case of observation noise alone, are determined by the intensity of the process noise and the characteristics of the system itself (cf.\@ \citep{bag2014} and Appendix for more details). We could see that our method (ELTO) shows smaller estimation errors compared with sDMD using delayed observation and eDMD, and was comparable to sDMD using arbitrary extension observations, even without the explicit extension.
As a total, both results show ELTO has a certain robustness to both observation noise and process noise, and it is capable of estimating eigenvalues of stochastic nonlinear system with smaller or comparable errors compared to the comparison DMD-based methods.

\section{Conclusions}
\label{sec:conclusion}

We formulated an operator-based latent Markov representation of a stochastic nonlinear dynamical system, where the stochastic evolution of the latent state embedded in a learnable RKHS is described with the corresponding transfer operator (ELTO), and developed a spectral method to learn this representation based on the theory of stochastic realization. We proposed the generalization of sequential state-estimation in stochastic nonlinear systems, and of operator-based eigen-mode decomposition of dynamics, for the representation. Finally, we showed several empirical results with synthetic and real-world data in sequential state-estimation and mode decomposition, demonstrating the superior performance of our model for these tasks.


\appendix




\renewcommand{\bibname}{References}
\renewcommand{\bibsection}{\subsubsection*{\bibname}}

%
%
\renewcommand{\thesection}{\Alph{section}}

\setlength{\textfloatsep}{10pt} 
\setlength{\floatsep}{10pt}     
\setlength{\intextsep}{10pt}    

\runningtitle{Learning Nonlinear Stochastic Dynamics with Embedded Latent Transfer Operators}


%

%

\onecolumn
\aistatstitle{Supplementary Material: \\Learning Nonlinear Stochastic Dynamics with \\Embedded Latent Transfer Operators}
\section{Proofs}
\label{app:proof}

\textbf{Proposition 1}

We denote the covariance matrices for the future and past respectively by 
\begin{align}
    (T_f)_{i,j} = {E}[u(\mathbf{y}(t+i-1))u(\mathbf{y}(t+j-1))]  \text{ and }  (T_p)_{i,j} = {E}[u(\mathbf{y}(t-i))u(\mathbf{y}(t-j))] (i, j = 1, 2, \dots),
\end{align}
Note that these are infinite-dimensional Toeplitz matrices (Toeplitz operators), and thus have the Cholesky factorizations, which we denote respectively by \( T_f = LL^\top \) and \( T_p = MM^\top \) \citep{CWS82}. Also from the assumption that \(\text{rank}(H) = r\), the SVD of \(L^{-1}HM^{-\top}\) is given by \( U\bm{S}V^\top \), where \( \bm{S} = \text{diag}(s_1, \dots, s_r) \). Note that the \(l\)-th canonical direction for \(\mathbf{y}_p(t)\) with respect to the CCA between \(\mathbf{y}_p(t)\) and \(\mathbf{y}_f(t)\) is given by the \(l\)-th column of \(V^\top M^{-1}u(\mathbf{y}_p(t))\), where \( u(\mathbf{y}_p(t)) := [u(\mathbf{y}(t-1)), u(\mathbf{y}(t-2)), \dots]^\top \). Then, the orthonormal projection of an element \(\xi_f = \sum_i c_i u(\mathbf{y}(t+i-1)) \in \mathbb{S}_t^+\) onto the past subspace \(\mathbb{S}_t^-\) is calculated as 
\begin{align}
    \mathcal{P}(\xi_f | \mathbb{S}_t^-) = c^\top H T_p^{-1} u(\mathbf{y}_p(t)) = c^\top LU \bm{S}^{1/2} \mathbf{x}(t),
\end{align}
where \( c := [c_1, c_2, \dots]^\top \) and \( \mathbf{x}(t) := \bm{S}^{1/2} V^\top M^{-1} u(\mathbf{y}_p(t)) \). We see that \( \mathbf{x}(t) \) can be a set of orthonormal bases of \( \mathbb{X}_t \).

\textbf{Theorem 1}

If we let \( \mathbf{x}_{t+1} = \bm{d}^\top \mathbf{x}(t+1) \in \mathbb{X}_{t+1} \), then from the definition of the orthonormal projection, we have
\begin{align}
    \mathcal{P}(\mathbf{x}_{t+1} | \mathbb{X}_t) = \bm{d}^\top E[\mathbf{x}(t+1) \mathbf{x}(t)^\top] E[\mathbf{x}(t)\mathbf{x}(t)^\top] \mathbf{x}(t) = \bm{d}^\top \tilde{\bm{A}} \mathbf{x}(t),
\end{align}
where \( \tilde{\bm{A}} \in \mathbb{R}^{r \times r} \). From the proof of Proposition 1, we note that \( \mathcal{Q} = LU \bm{S}^{1/2} \) and 
\begin{align}
    \mathbf{x}(t) = \bm{S}^{1/2} V^\top M^{-1} u(\mathbf{y}_p(t)). \quad 
\end{align}

Also, if we define \( \mathcal{Z} := \bm{S}^{1/2} V^\top M^\top \), then from the definition, the Hankel operator has the decomposition \( H = \mathcal{Q}\mathcal{Z} \) due to \( \text{rank}(H) = r \), and we have \( \mathbf{x}(t) = \mathcal{Z}T_p^{-1} u(\mathbf{y}_p(t)) \). Moreover, we have \( H^\uparrow = \mathcal{Q}^\uparrow\mathcal{Z} = \mathcal{Q} \mathcal{Z}^\leftarrow \) (see, for example, Theorem 6.1 in  \citep{Kat05}). Here, the operators $\bullet^{\leftarrow}$ and $\bullet^{\uparrow}$, acting on the block matrix. correspond to the operations of removing the first block column and the first block row, respectively. Therefore, we have 
\begin{align}
    \tilde{\bm{A}} &= \bm{S}^{1/2} V^\top M^{-1} E[u(\mathbf{y}_p(t+1)) u(\mathbf{y}_p(t))^\top] T_p^{-1} M V \bm{S}^{-1/2} \\
&= \bm{S}^{1/2} V^\top M^{-1} M (M^\top)^\leftarrow (M M^\top)^{-1} M V \bm{S}^{-1/2} \\
&= \bm{S}^{1/2} V^\top (M^\top)^\leftarrow (M^\top)^{-1} V \bm{S}^{-1/2} \\
&= \mathcal{Z}^\leftarrow \mathcal{Z}^\dagger  = \mathcal{Q}^\dagger  \mathcal{Q}^\uparrow (= \bm{A}).
\end{align}

Moreover, from the definition and Eq.(4), it is obvious that \( \mathbf{w}(t) \perp \mathbb{S}_t^- \).
\newpage
\textbf{Corollary 1}

Let $\textbf{x}(t)$ and $\textbf{x}(t+1)$ be random variables with corresponding marginal distributions $\mathbb{P}(\textbf{x}(t))$ and $\mathbb{P}(\textbf{x}(t+1))$, and $\bm{x}_t$ and $\bm{x}_{t+1}$ are respectively realizations of these random variables. Let $\mathbb{H}$ be the RKHS associated with a kernel function $k$. As described in Section~3 of the main text, the kernel Koopman operator $\mathcal{K}_k$ maps an observation function $g$ to the expectation ${E}[g(\textbf{x}(t+1))|\textbf{x}(t)=\cdot]$ under the assumption that these functions are in $\mathbb{H}$, which is formulated as follows
\begin{align*}
(\mathcal{K}_k g)(\bm{x}_{t}) = \int p_{\text{tr}}(\bm{x}_{t+1}|\bm{x}_t) \int p_{\text{ob}}(\bm{y}|\bm{x}_{t+1})\bm{y}d\bm{y}d\bm{x}_{t+1} = {E}[g(\textbf{x}(t+1))|\textbf{x}(t) = \bm{x}_t]
\end{align*}%
where $\bm{y}$ is an element of any measurable set over which the integral is taken. 
Here, from Section~3 of \citep{Fukumizu2004DimensionalityRF}, if $E[g(\textbf{x}(t+1))|\textbf{x}(t) = \cdot] \in \mathbb{H}$ for all observable $g\in\mathbb{H}$, it satisfies that
\begin{align*}%
\mathcal{C}_{\textbf{x}(t)}{E}[g(\textbf{x}(t+1))|\textbf{x}(t) = \cdot] =  \mathcal{C}_{\textbf{x}(t)\textbf{x}(t+1)} g,
\end{align*}
where $\mathcal{C}_{\textbf{x}(t)}, \mathcal{C}_{\textbf{x}(t)\textbf{x}(t+1)}$ are the (cross) covariance operators. Note that based on the assumptions in this text, $\textbf{x}(t)$ and $\textbf{x}(t+1)$ are elements in the same space. As a result, we have  $\mathcal{C}_{\textbf{x}(t)}\mathcal{K}_k g =  \mathcal{C}_{\textbf{x}(t)\textbf{x}(t+1)}g$  and the equation of the corollary follows. $\Box$

Note that the regularity of $\mathcal{C}_{\textbf{x}(t)}$ remains a subject of discussion because the assumption, i.e., $E[g(\textbf{x}(t+1))|\textbf{x}(t) = \cdot] \in \mathbb{H}$ for all observable $g\in\mathbb{H}$, may not be hold in general. As note in \citep{FSG13} and \citep{GKN19}, A common approach to alleviate this problem is to consider the regularized variant $\mathcal{C}_{\textbf{x}(t)} + \epsilon \mathcal{I}$ ($\mathcal{I}$ is the identity operator) to ensure the regularity.

\section{Algorithms}
\subsection{Estimating Embedded Latent Transfer Operators (ELTOs) with Spectral Learning}

In this section, we describe the algorithm to estimate the state process and compute ELTOs and EOOs, which is shown in Algorithm~\ref{alg1}. Here, unless explicitly stated otherwise, all notations are based on those defined in the main text. The procedure is divided into two parts. That is, we first outline the supplementation of covariance matrices and the computation of singular-value decomposition. Then, we estimate the empirical values of $\hat{\mathcal{T}}_e$ and $\hat{\mathcal{O}}_e$, based on the trained $\bm{w}$. Please note that all assumptions are consistent with those outlined in Section 4 of the main text.\par
\begin{algorithm}[t]
    \caption{Estimating Embedded Latent Transfer Operators with Spectral Learning}
    \label{alg1}
    \KwIn{dataset $\mathbf{Y}=[\bm{y}_0,\bm{y}_1,...,\bm{y}_T]$, window size $h>2$, max epoch size $L$, form of kernel function $k(\cdot, \cdot)$}
    \vspace{5pt}
    \textit{Compute the Gram matrix using the kernel function:}\\
    \Indp
    $(\mathbf{G})_{(i,j)} =(\Phi^\top  \Phi)_{(i,j)} = k  (\bm{y}_i, \bm{y}_j) $\\
    \vspace{3pt}
    where $i,j = 0,\dots, N$, $N=T-2h+2$ and $\Phi= [\phi(\bm{y}_0),\phi(\bm{y}_1),...,\phi(\bm{y}_N) ] $\\
    \Indm
    \vspace{5pt}

    \For{$l=1:L$}{
        \vspace{3pt}
        \textit{Estimate self-covariance matrices and cross-covariance matrix for $u(\rvy_f(t))$ and $u(\rvy_p(t))$:}\\
        \vspace{3pt}
        \Indp
        $(\mathbf{C}_{pp})_{(i_1,i_2)} = \bm{w}^\top \mathbf{G}_{S,i_1+1:i_1+N} \mathbf{Q}_N \mathbf{G}_{i_2+1:i_2+N, S}  \bm{w}$, \\
        $(\mathbf{C}_{ff})_{(j_1,j_2)} = \bm{w}^\top \mathbf{G}_{S,j_1+1:j_1+N} \mathbf{Q}_N \mathbf{G}_{j_2+1:j_2+N, S}  \bm{w}$, \\   
        $(\mathbf{C}_{fp})_{(j_1,i_2)} = \bm{w}^\top \mathbf{G}_{S,j_1+1:j_1+N} \mathbf{Q}_N \mathbf{G}_{i_2+1:i_2+N, S}  \bm{w}$ $\quad$( $i_1, i_2, j_1, j_2 = 1, \dots, h$ )\\
        for any sampled subset $S =\{e_1, \dots, e_{|S|}\}\subseteq  \{ 0,1,...,T \}$, \\
        the sub-matrix $\mathbf{G}_{\bullet 1,\bullet 2}$ consisting of $\bullet 1 / \bullet 2$ columns and rows of $\mathbf{G}$,\\
        and the normalization matrix $\mathbf{Q}_N:=\mathbf{I}_N-(1/N)\bm{1}_N\bm{1}_N^\top$.\\
        \vspace{7pt}
        \Indm
        \textit{Calculate canonical correlation analysis with SVD :}\\
        \Indp
        $(\mathbf{U}_{ff}, \mathbf{S}_{ff}, \mathbf{V}_{ff}) = \text{SVD}(\mathbf{C}_{ff})$, \hspace{2pt} $(\mathbf{U}_{pp}, \mathbf{S}_{pp}, \mathbf{V}_{pp}) = \text{SVD}(\mathbf{C}_{pp})$\\
        $\mathbf{S}_f = \sqrt{\text{diag}(\mathbf{S}_{ff})}$, \hspace{2pt}  $\mathbf{S}_p = \sqrt{\text{diag}(\mathbf{S}_{pp})}$\\
        $\mathbf{L} = \mathbf{U}_{ff} \cdot \mathbf{S}_f \cdot \mathbf{V}_{ff}$, \hspace{2pt} $\mathbf{M} = \mathbf{U}_{pp} \cdot \mathbf{S}_p \cdot \mathbf{V}_{pp}$\\
        $\mathbf{L}^{-1} = \mathbf{V}_{ff}^\top \cdot (\mathbf{S}_f)^{-1} \cdot \mathbf{U}_{ff}^\top$, \hspace{2pt} $\mathbf{M}^{-1} = \mathbf{V}_{pp}^\top \cdot (\mathbf{S}_p)^{-1} \cdot \mathbf{U}_{pp}^\top$\\
        $(\mathbf{U}, \mathbf{S}, \mathbf{V}) = \text{SVD}(\mathbf{L}^{-1} \cdot \mathbf{C}_{fp} \cdot (\mathbf{M}^{-1})^\top)$\\
        \Indm
        \vspace{7pt}
        \textit{Optimizing $\bm{w}$ by the loss function:}\\
        \Indp
        $L(\bm{w}) = -\sum_{i=1}^h diag(\mathbf{S})_i$\\
        Update $\bm{w}$ using Adam with $L(\bm{w})$ \\
        \vspace{3pt}
        \Indm
        
    }
    \textbf{return} trained $\bm{w}$\\
    \vspace{3pt}
    
    \textit{Estimate empirical state process:}\\
    \Indp
    Set subsequences: $\bm{y}_{p,n} = [\bm{y}_{n+h-2}^\top,...,\bm{y}_{n-1}^\top ]^\top$ \\
    $\bm{x}_n:= \mathbf{B} \Phi_{p,n}^\top \Phi_S \bm{w};$\\
    where $\Phi_S = [\phi(\bm{y}_{e_1}),...,\phi(\bm{y}_{e_{|S|}})]$, $\Phi_{p,n} = [\phi(\bm{y}_h),...,\phi(\bm{y}_{h+N-1})]$\\
    
    \Indm
    \Return{the state process $\{\bm{x}_n\}_{n=0,\dots,N}$} \\
    \vspace{5pt}
    \textbf{Compute operators} $\hat{\mathcal{T}}_e$ \textbf{and} $\hat{\mathcal{O}}_e$: \\
    \Indp
    Set Feature and Gram matrices:\\
    \Indp
    $\Psi = [\phi(\bm{x}_0),\dots, \phi(\bm{x}_N)]$, $\Psi_1 = \Psi_{:,1:N}$, $\Psi_2 = \Psi_{:,2:N+1}$, $\mathbf{G}_x = \Psi_1^\top \Psi_1$, $\mathbf{G}_{yx} = \Phi^\top \Psi$, $\epsilon_t,\epsilon_o > 0$\\
    \Indm
    Formulate the empirical estimation of the operator: \\
    \Indp
    $\hat{\mathcal{T}}_e = \Psi_2 \Psi_1^\top ( \Psi_1 \Psi_1^\top + \epsilon \mathcal{I}) ^ {-1} =
    \Psi_2(\mathbf{G}_x + \epsilon_t \mathbf{I}_{N-1})\Psi_1^\top$,$\quad$$\hat{\mathcal{O}}_e = \Phi \Psi^\top (\Psi \Psi^\top + \epsilon \mathcal{I})^{-1}=
    \Phi(\mathbf{G}_{yx} + \epsilon_o \mathbf{I}_{N-1})\Psi^\top$\\
    \Indm
    \Indm
    
\end{algorithm}
Note that information about the time evolution of the system can be obtained by directly analyzing the estimated state system. The numerical experiments related to this correspond to Chapter 6 in the main text and Appendix C.4 and C.5.

\subsection{Sequential-state estimation with ELTOs}
We present the algorithm for the sequential state estimation described in Section 5 (Algorithm~\ref{alg4}). Here, we also list some initializations for belief states, together with the update loop of filtering. Note that Algorithm 2 performs updates of the empirical approximation of operators in a finite-dimensional space, which is based on \citep{GKN19}. 

\begin{algorithm}[t]
    \caption{Sequential state estimation with ELTOs}
    \label{alg4}
    \KwIn{$\mathbf{Y} = [\bm{y}_0, \bm{y}_1, \dots, \bm{y}_T]$, regularization parameters $\epsilon_t, \epsilon_o, \epsilon_q$, form of kernel function $k(\cdot, \cdot)$}
    \textit{Compute the state process as in Algorithm 1}:  $\{\bm{x}_n\}_{n=0}^N$ \\
    \textit{Compute following Gram matrices using Algorithm 1:} \\
    \Indp
    $\mathbf{G}_y =[\phi(\bm{y}_1), \dots, \phi(\bm{y}_N)]^\top  [\phi(\bm{y}_1), \dots, \phi(\bm{y}_N)] $, $\mathbf{G}_x = \Psi_2^\top \Psi_2$, $\mathbf{G}_{\tilde{x}}=\Psi_1^\top \Psi_1$, $\mathbf{G}_{\tilde{{x}}x}=\Psi_1\Psi_2$\\
    \Indm
    \textit{Initialization:} \\
    \Indp
    $(\mathbf{K}_0)_{i,j} = k_x(\bm{x}_i, \bm{u}_j)$;\\
    where $i=1,\dots,N$ and $\bm{u}_j$ is a sample from multivariate Uniform distribution in the range (0,1) for all $j$\\
    \Indm
    \textit{Compute the matrices based on ELTO}: \\
    \Indp
    $\mathbf{T} = (\mathbf{G}_{\tilde{x}} + \epsilon_t \mathbf{I}_{N})^{-1} \mathbf{G}_{\tilde{{x}}x}$, $\mathbf{O} = (\mathbf{G}_x + \epsilon_o \mathbf{I}_N)^{-1} \mathbf{G}_x$\\
    \Indm
    \textit{Compute initial embeddings}:$\quad\mathbf{C}_0 = (\mathbf{G}_x + \epsilon_o \mathbf{I}_N)^{-1} \mathbf{K}_0$\\
    \textit{Compute mean} $\bm{m}_0$ \textit{and variance} $\mathbf{S}_0$ \textit{}over the columns of $\mathbf{C}_0$\\

    \For{$t < T$}{
        \If{new observation $\bm{y}_t$ available}{
            Compute the matrix based on Kalman gain operator:\\
            \vspace{3pt}
            \Indp
            $\mathbf{Q}_t = \mathbf{S}_t^-\mathbf{O}^\top(\mathbf{G}_y\mathbf{O}\mathbf{S}_t^-\mathbf{O}^\top + \epsilon_q \mathbf{I}_N)^{-1}$\\
            \Indm
            Innovation update:\\
            \Indp
            $\bm{m}_t^+ = \bm{m}_t^- + \mathbf{Q}_t ([k(\bm{y}_1, \bm{y}_t), \dots, k(\bm{y}_N, \bm{y}_t)]^\top - \mathbf{G}_y\mathbf{O} \bm{m}_t^-)$\\
            $\mathbf{S}_t^+ = \mathbf{S}_t^- - \mathbf{Q}_t \mathbf{G}_y \mathbf{O}\mathbf{S}_t^-$\vspace{-0.5em}\\
            \Indm
            \vspace{10pt}
        }
        \hspace{0.5em}
        Prediction update:\hspace{0.5em} \\
        \Indp
        $\bm{m}_{t+1}^- = \mathbf{T}\bm{m}_t^+$,\hspace{0.5em} \\
        $\mathbf{S}_{t+1}^- = \mathbf{T}\mathbf{S}_t^+ \mathbf{T}^\top + \frac{1}{N}\left((\mathbf{G}_{\tilde{x}} + \epsilon_t \mathbf{I}_N)^{-1} \mathbf{G}_{\tilde{x}} - \mathbf{I}_N\right)\left((\mathbf{G}_{\tilde{x}} + \epsilon_t \mathbf{I}_N)^{-1} \mathbf{G}_{\tilde{x}} - \mathbf{I}_N\right)^\top$;\\
        \Indm
        \vspace{3pt}
        Project into original observation space:\hspace{0.5em}\\
        \vspace{3pt}
        \Indp
        $\hat{\bm\eta}_t = [\bm{y}_1, \dots, \bm{y}_N] (\mathbf{G}_{\tilde{x}} + \epsilon_o \mathbf{I}_N)^{-1} \mathbf{G}_x \bm{m}_t^+ $,\\
        \vspace{3pt}
        $\hat{\bm{\Sigma}}_{t} = [\bm{y}_1, \dots, \bm{y}_N] (\mathbf{G}_{\tilde{x}} + \epsilon_o \mathbf{I}_N)^{-1}\mathbf{G}_x \mathbf{S}_t^+ \mathbf{G}_x^\top \left((\mathbf{G}_{\tilde{x}} + \epsilon_o \mathbf{I}_N)^{-1}\right)^\top\mathbf{Y}^\top$\vspace{-0.5em} \\
        \Indm
        \vspace{10pt}
    }
\end{algorithm}

\section{Experimental settings}
In this chapter, we explain the detailed settings for the numerical experiments described in the main text.

\subsection{Pendulum}
We generate the pendulum dataset as described in Section 5.3.1 of \citep{GKN19}. The simulation was performed using the code in \texttt{https://github.com/gregorgebhardt/pyKKR} .


\textbf{Simulation setting}
\begin{itemize}[itemsep=0.1pt, topsep=0.1pt]
    \item Initial conditions: angle $q_0 \sim \mathcal{U}(0.1\pi, 0.4\pi)$, \quad velocity $\dot{q}_0 \sim \mathcal{U}(-0.25\pi, 0.25\pi)\quad$  ($\mathcal{U}$: Uniform distribution)
    \item Frequency: 10,000 Hz
    \item Simulation number : 1,500 for training;  500 for validation
    \item Process noise: $p_t \sim \mathcal{N}(0, 0.1)\quad$  ($\mathcal{N}$: Normal distribution)
    \item Observation noise: $o_t \sim \mathcal{N}(q_t, 0.01)$ \qquad\qquad\qquad 
\end{itemize}

\textbf{Parameters for ELTO (our method)}
\begin{itemize}[itemsep=0.1pt, topsep=0.1pt]
    \item Kernel function for Gram matrix : RBF kernel with $\sigma$ = 1
    \item Training using Adam with learning rate $10^{-3}$
\end{itemize}

Note that $\mathcal{U}$ and $\mathcal{N}$ are Uniform and Gaussian distribution, respectively. In this experiment, we used all sequences for $\Phi_S$ and did not truncate any singular value for estimating state variables $\bm{x}_n$. Additionally, we used an experimental setup with epoch = 200 and window size = 5 for comparison. In the following, we provide additional analysis on the impact of different combinations of settings on the experimental data. In this case, we evaluated for MSE between the observation and the prediction after the pre-image step (\cite{GKN19}; Sec.5.1.2). The result is Table~\ref{fig:performance-table}.
\begin{table}[H]
\centering
\begin{tabular}{cccc}
\hline
Model:ELTO & optimal MSE($\times 10^{-4}mm$) & train time  & avg eval time  \\ \hline
epoch=200, window size=5  & 1.0121 & 4.22s & 40$\pm$1 s \\
epoch=200, window size=8  & 1.1089 & 9.88s & 40$\pm$1 s \\
epoch=500, window size=5  & 1.0260 & 12.03s & 40$\pm$1 s\\ 
epoch=500, window size=8  & 1.1226 & 24.34s & 40$\pm$1 s\\
\hline
\end{tabular}
\caption{The performance of proposal method under different settings in Pendulum experiment}
\label{fig:performance-table}
\end{table}

Note that for a sequence of length 1500, with the same window size (i.e., specifying the model's preference for short-term or long-term dependencies), around 200 epochs are sufficient for convergence. Additional training epochs do not lead to better results and can even lead to overfitting. Furthermore, since the evaluation using CMA-ES \citep{hansen2006cma} is independent of this, the choice of parameters does not affect the evaluation speed.
This comparison presented was conducted on an NVIDIA 4060 Ti GPU.

We also consider the performance under different levels of different noise, where $n_p$ and $n_o$ stand for the standard deviation (std) of the generated process noise and observation noise, and the result, which is shown in Table ~\ref{fig:additional-exp-results},  is the converged ones after 200 epoch.

\begin{table}[H]
\centering
\begin{tabular}{cccc}
\hline
\multirow{2}{*}{Noise setting} & \multicolumn{3}{c}{MSE($\times 10^{-3}$)
} \\ \cline{2-4}
& SubKKR & KKR & ELTO(ours) \\ \hline
$n_p$=0.1, $n_o$=0.01 & 2.7574 & 1.1185 & \textbf{0.1010} \\ 
$n_p$=0.2, $n_o$=0.01 & 3.9321 & 3.4061 & \textbf{0.2325} \\ 
$n_p$=0.1, $n_o$=0.1 & 13.192 & 15.797 & \textbf{11.278} \\ 
\hline
\end{tabular}
\label{fig:additional-exp-results}
\caption{Model performance under different noise settings}
\end{table}

\subsection{HuMoD Dataset}

In this experiment, we use an open database modeling and simulation of human motion dynamics by \citep{wojtusch2015humod}. As described in the original paper, the participants performed various activities, including steady walking, steady running, variable-speed running, and other motions. The 36 markers recorded the three-dimensional positions of various body parts, while 14 additional sensors captured muscle activity and other related information. In our experiment, we focused on the height (markerY) of the 8th and 12th thoracic vertebrae (T8/T12) for analysis. Here, we will only provide the information necessary for this experiment, but for more detailed discussion on the sensors, please refer to \citep{wojtusch2015humod}.

\textbf{data setting for this information}
\begin{itemize}[itemsep=0.1pt, topsep=0.1pt]
    \item Frequency: 50Hz which is downsampled from original data 500Hz
    \item Simulation number: 500 for training; 200 for validation
\end{itemize}

\textbf{Parameter setting for ELTO (our method)}
\begin{itemize}[itemsep=0.1pt, topsep=0.1pt]
    \item Kernel function for Gram matrix : RBF kernel with $\sigma$ = 1
    \item Training using Adam with learning rate $10^{-3}$
\end{itemize}

Note that, as in the above experiment,  we used all sequences for $\Phi_S$ and did not truncate any singular value for estimating state variables $\bm{x}_n$. Additionally, we optimized the regularization term using CMA-ES \citep{hansen2006cma}.
We run this experiment on Tesla V100S-PCIE-32GB.


\subsection{Quad-link}

In this experiment, We generate the dataset using the simulator in \texttt{https://github.com/ALRhub/rkn\_share}, as the comparative method. Regarding the simulation, we will describe the information on the parameters necessary for the experiment here, which mostly follow the ones in \citep{BPG+19}:

\textbf{Simulation setting}
\begin{itemize}[itemsep=0.1pt, topsep=0.1pt]
    \item Initial angles: sampling from $\mathcal{U}(-\pi, \pi)$
    \item Friction coefficient: 0.1
    \item time step: $dt = 10^{-4}$, and data is collected every 0.05 seconds.
    \item Simulation number: 1,200 for training ELTO; 300 for training decoder; 200 for validation
    \item Noise parameter: $r = 0.2$  \quad (which means the temporary correlation of noises)
\end{itemize}
The details of the training will be explained later. Note that the other parameter settings are based on the defaults of the aforementioned simulator.


\textbf{Prameters for ELTO (our method)}

In this experiment, we used the deep kernel (which can be interpreted as an \textbf{encoder}), and decoders to reconstruct the original data from the features. We first trained the deep kernel with full dataset, and retrain the decoders using a subset that we randomly subsample. The structure of these network was designed as follows:

\textbf{Encoder:}
\begin{itemize}[itemsep=0.1pt, topsep=0.1pt]
    \item Convolution 1: 12 $5 \times 5$ filter, ReLU $2 \times 2$ max-pooling with $2 \times 2$ stride.
    \item Convolution 2: 12 $3 \times 3$ filter with $2 \times 2$ stride, ReLU $2 \times 2$ max-pooling with $2 \times 2$ stride.
    \item Fully Connected layer: ReLU 200 (the output dimension is the same as the comparative method)
\end{itemize}

\textbf{Decoder:}
\begin{itemize}[itemsep=0.1pt, topsep=0.1pt]
    \item Mean: Linear: $200 \rightarrow 64$, Tanh, $64 \rightarrow 8$, Tanh.
    \item Variance: Linear: 300 $\rightarrow$ 64, Tanh, Linear: 64 $\rightarrow $8, Tanh.
\end{itemize}

\textbf{(Other setting)}

\begin{itemize}[itemsep=0.1pt, topsep=0.1pt]
    \item Kernel function for Gram matrix : deep RBF kernel with $\sigma = 1/200$
    \item Adopted data for $\Phi_S$ : all data
    \item Both training stages used the Adam optimizer with a learning rate $10^{-3}$
    \item Both training epochs: 200 
\end{itemize}

We performed a second round of training for the encoder and decoder using image-format input (with $48 \times 48$ dimension) and preimage-output (with 8 dimension), as described in Section 5.3 of the main text. The Gaussian negative log-likelihood was used as the training loss function, while the mean squared error was used for evaluation.
We run this experiment on Tesla V100S-PCIE-32GB.

\subsection{Van Der Pol oscillator with the observation noise}

 We generate the simulation dataset as described in Section 6.1.1 in the main text. The VDP oscillator forms a limit cycle in the phase space, determined by the parameter $\mu$. The details of the experiment setting as follows :

\textbf{Simulation setting}
\begin{itemize}[itemsep=0.1pt, topsep=0.1pt]
    \item Time step: $dt = 0.1$
    \item Initial condition: $(x, y) = (2.0, 0.0)$
    \item Simulation number: 3,000 for training, 500 for validation
    \item Numerical solution: fourth-order Runge-Kutta method
\end{itemize}

\textbf{Parameters for the asymptotic system}
\begin{itemize}[itemsep=0.1pt, topsep=0.1pt]
    \item Nonlinear term of $y$ : $\mu = 2$
\end{itemize}
 
\textbf{Parameters for ELTO (our method)}
\begin{itemize}[itemsep=0.1pt, topsep=0.1pt]
    \item Kernel function for Gram matrix : RBF kernel with $\sigma = 0.01$ 
    \item Adopted number of singular values (correlation coefficients) : 9
    \item Adopted data for $\Phi_S$ : the last 600 data
    \item Training for 50 epochs using Adam with learning rate $10^{-5}$.
\end{itemize}
In nonlinear oscillators such as the VDP oscillator, higher-order harmonics appear in addition to the fundamental frequency. Therefore, in this experiment, the period of the numerically computed limit cycle is taken as the fundamental angular frequency, and the eigenvalues corresponding to integer multiples of this angular frequency are expected to appear. In this experiment, we set the fundamental angular frequency to $\omega = 0.823498$ based on a simulation without observation noise.

The eigenvalues were estimated by calculating Extend DMD \citep{WKR15} based on the estimated state system. In this calculation, the same kernel used for the state variable estimation was applied. On the other hand, for the comparison methods, DMD and Subspace DMD \citep{TKY17a}, the eigenvalues were estimated using a Hankel matrix accumulated from 30 observations as input.

The experiment was conducted by increasing the intensity of the observation noise, with the variance ranging from 0.01 to 0.2, in increments of 0.01. The absolute error between the expected eigenvalues and the estimated eigenvalues was used to evaluate the estimation error. For each experiment, 50 trials were conducted, and the mean and standard deviation were calculated. Note that this experiment was conducted on an NVIDIA A100 40GB GPU.

\subsection{Stuart Landau oscillator with the process noise}

As above experiment,  We generate the simulation dataset as described in Section 6.1.2 in the main text. The SL oscillator also forms a limit cycle in the phase space. On the other hand, if there is no process noise, the limit cycle takes a circular shape, and its period is determined by the parameters , which is calculated by $2\pi / (\gamma - \beta\mu)$. However, when process noise is present, the trajectory moves along the circle in phase space, oscillating around the circumference. The estimated eigenvalues also differ from those in the absence of noise. In this section, we will explain only the basic facts. For theoretical details, please refer to \citep{bag2014}.

The details of the experiment setting as follows :

\textbf{Simulation setting}
\begin{itemize}[itemsep=0.1pt, topsep=0.1pt]
    \item Time step: $dt = 0.1$
    \item Initial condition: $(r, \theta) = (0.1, 0.0)$
    \item Simulation number: 3,000 for training, 500 for validation
    \item Numerical solution: fourth-order Runge-Kutta method
\end{itemize}

\textbf{Parameters for the asymptotic system}
\begin{itemize}[itemsep=0.1pt, topsep=0.1pt]
    \item ( for $r$ ) : $\mu =  1.0$
    \item ( for $\theta$ ) : $\gamma = 0.9, \ \beta = 0.3$
\end{itemize}

\textbf{Parameters for ELTO (our method)}
\begin{itemize}[itemsep=0.1pt, topsep=0.1pt]
    \item Kernel function for Gram matrix : RBF kernel with $\sigma = 0.01$ 
    \item Adopted number of singular values (correlation coefficients) : 13
    \item Adopted data for $\Phi_S$ : the last 600 data
    \item Training for 50 epochs using Adam with a learning rate $10^{-5}$.
\end{itemize}

Regarding the estimated eigenvalues, \citep{bag2014} demonstrated that when weak process noise is applied to the SL oscillator, the $m$-th eigenvalue (which is related to $m$-th harmonics) in the continuous system can be approximately calculated using the following relation:
\begin{align*}
    s(\omega, \epsilon) = i m\omega - \frac{\epsilon}{2} \kappa m^2 \omega^2 + \mathcal{O}(\epsilon^2),
\end{align*}
where $\omega$ is the fundamental frequency, $\epsilon$ is the variance of process noise , and $\kappa$ is the constant following the dynamics without process noise \citep{bag2014}. In this experiment, we set $\omega = 0.6$, $\kappa = 3.0$.

The method for estimating eigenvalues in the proposed approach is the same as in the previous experiment.For the observation, we used the exponential of the angle $\theta$ as the value $e^{i \theta}$. On the other hand, for comparison methods, we used Extend DMD \citep{WKR15} and Subspace DMD based on the observation function 
\begin{align*}
    g(\theta) = (e^{-10i\theta},\ e^{-9i\theta}, \ \dots\ , e^{9i\theta},e^{10i\theta})
\end{align*}
from \citep{TKY17a}, as well as Subspace DMD, which accumulates 30 observations.

The experiment was conducted by increasing the intensity of the observation noise, with the variance ranging from 0.01 to 0.09, in increments of 0.01.The method for calculating the estimation error, its mean, and standard error is the same as in the above experiment. Note that this experiment was conducted on an NVIDIA A100 40GB GPU.

\begin{thebibliography}{}

\bibitem[Akaike, 1975]{Aka75}
Akaike, H. (1975).
\newblock Markovian representation of stochastic processes by canonical variables.
\newblock {\em SIAM Journal on Control}, 13(1):162--173.

\bibitem[Arbabi and Mezi\'{c}, 2017]{AM17}
Arbabi, H. and Mezi\'{c}, I. (2017).
\newblock Ergodic theory, dynamic mode decomposition, and computation of spectral properties of the {K}oopman operator.
\newblock {\em SIAM Journal on Applied Dynamical Systems}, 16(4):2096--2126.

\bibitem[Azencot et~al., 2020]{AELM20}
Azencot, O., Erichson, B., Lin, V., and Mahoney, M.~W. (2020).
\newblock Forecasting sequential data using consistent koopman autoencoders.
\newblock In {\em Proc.\@ of the 37th Int'l Conf. on Machine Learning (ICML'20)}, pages 475--485.

\bibitem[Bagheri, 2014]{bag2014}
Bagheri, S. (2014).
\newblock Effects of weak noise on oscillating flows: Linking quality factor, floquet modes, and koopman spectrum.
\newblock {\em Physics of Fluids}, 26(9):094104.

\bibitem[Baker, 1973]{baker1973}
Baker, C.~R. (1973).
\newblock Joint measures and cross-covariance operators.
\newblock {\em Transactions of the American Mathematical Society}, 186:273--289.

\bibitem[Becker et~al., 2019]{BPG+19}
Becker, P., Pandya, H., Gebhardt, G., Zhao, C., Taylor, C., and Neumann, G. (2019).
\newblock Recurrent kalman networks: Factorized inference in high-dimensional deep feature spaces.
\newblock In {\em Proc.\@ of the 36th Int'l Conf.\@ on Machine Learning}, volume PMLR 97, pages 544--552.

\bibitem[Brunton et~al., 2022]{BBKK22}
Brunton, S.~L., Budisi\'{c}, M., Kaiser, E., and Kutz, J.~N. (2022).
\newblock Modern koopman theory for dynamical systems.
\newblock {\em SIAM REVIEW}, 64(2):229--340.

\bibitem[Budisi\'{c} et~al., 2012]{BMM12}
Budisi\'{c}, M., Mohr, R., and Mezi\'{c}, I. (2012).
\newblock Applied koopmanism.
\newblock {\em Chaos}, 22:047510.

\bibitem[Chen et~al., 2012]{CTR12}
Chen, K., Tu, J., and Rowley, C. (2012).
\newblock Variants of dynamic mode decomposition: Boundary condition, koopman, and fourier analyses.
\newblock {\em Journal of Nonlinear Science}, 22:887--915.

\bibitem[Chiuso, 2005]{Chi05}
Chiuso, A. (2005).
\newblock On the relation between {CCA} and predictor-based subspace identification.
\newblock {\em IEEE Transactions on Automatic Control}, 52(10):1795--1812.

\bibitem[Chui et~al., 1982]{CWS82}
Chui, C., Ward, J., and Smith, P. (1982).
\newblock Cholesky factorization of positive definite bi-infinite matrices.
\newblock {\em Numerical Functional Analysis and Optimization}, 5(1):1--20.

\bibitem[Dellnitz et~al., 2000]{DFJ00}
Dellnitz, M., Froyland, G., and Junge, O. (2000).
\newblock {\em The algorithms behind {GAIO} -- {S}et oriented numerical methods for dynamical systems}, pages 145--174.
\newblock Springer.

\bibitem[Desai et~al., 1985]{DPK85}
Desai, U., Pal, D., and Kirkpatrick, R. (1985).
\newblock A realization approach to stochastic model reduction.
\newblock {\em International Journal of Control}, 42(4):821--838.

\bibitem[Eubank and Hsing, 2008]{EH08}
Eubank, R. and Hsing, T. (2008).
\newblock Canonical correlation for stochastic processes.
\newblock {\em Stochastic Processes and their Applications}, 118(9):1634--1661.

\bibitem[Froyland and Padberg, 2009]{FP09}
Froyland, G. and Padberg, K. (2009).
\newblock Almost-invariant sets and invariant manifolds connecting probabilistic and geometric descriptions of coherent structures in flows.
\newblock {\em Physica D}, 238:1507.

\bibitem[Fukumizu et~al., 2004a]{FBJ04}
Fukumizu, K., Bach, F., and Jordan, M. (2004a).
\newblock Dimensionality reduction for supervised learning with reproducing kernel {H}ilbert spaces.
\newblock {\em Journal of Machine Learning Research}, 5:73--99.

\bibitem[Fukumizu et~al., 2004b]{Fukumizu2004DimensionalityRF}
Fukumizu, K., Bach, F.~R., and Jordan, M.~I. (2004b).
\newblock Dimensionality reduction for supervised learning with reproducing kernel hilbert spaces.
\newblock {\em J. Mach. Learn. Res.}, 5:73--99.

\bibitem[Fukumizu et~al., 2013]{FSG13}
Fukumizu, K., Song, L., and Gretton, A. (2013).
\newblock Kernel {B}ayes' rule:\@ {B}ayesian inference with positive definite kernels.
\newblock {\em Journal of Machine Learning Research}, 14:3753--3783.

\bibitem[Gebhardt et~al., 2019]{GKN19}
Gebhardt, G., Kupcsik, A., and Neumann, G. (2019).
\newblock The kernel kalman rule: {E}fficient nonparametric inference by recursive least-squares and subspace projections.
\newblock {\em Machine Learning}, 108:2113--2157.

\bibitem[Ghahramani and Hinton, 1996]{GH96}
Ghahramani, Z. and Hinton, G. (1996).
\newblock Parameter estimation for linear dynamical systems.
\newblock Technical report, Dept.\@ Comp.\@ Sci.\@, Univ.\@ Toronto.

\bibitem[Ghahramani and Hinton, 2000]{GH00}
Ghahramani, Z. and Hinton, G.~E. (2000).
\newblock Variational learning for switching state-space models.
\newblock {\em Neural Computation}, 12(4):831–864.

\bibitem[Ghahramani and Roweis, 1998]{GR19}
Ghahramani, Z. and Roweis, S.~T. (1998).
\newblock Learning nonlinear dynamical systems using an {EM} algorithm.
\newblock In {\em Advances in Neural Information Processing Systems 11}, page 431–437.

\bibitem[Gordon et~al., 1993]{GSS93}
Gordon, N., Salmond, D., and Smith, A. (1993).
\newblock Novel approach to nonlinear/non-gaussian bayesian state estimation.
\newblock In {\em IEE Proc.\@ F - Radar and Signal Processing}, volume 140, pages 107--113.

\bibitem[Grewal and Andrews, 2010]{GA10}
Grewal, M. and Andrews, A. (2010).
\newblock Application of {K}alman filtering in aerospace 1960 to the present.
\newblock {\em IEEE Control System Magazine}, 30:69--78.

\bibitem[Hannan and Deistler, 1988]{HD88}
Hannan, E. and Deistler, M. (1988).
\newblock {\em The Statistical Theory of Linear Systems}.
\newblock Wiley.

\bibitem[Hansen, 2006]{hansen2006cma}
Hansen, N. (2006).
\newblock The {CMA} evolution strategy: {A} comparing review.
\newblock {\em Towards a new evolutionary computation: Advances in the estimation of distribution algorithms}, pages 75--102.

\bibitem[Hsu et~al., 2009]{HKZ09}
Hsu, D., Kakade, S., and Zhang, T. (2009).
\newblock A spectral algorithm for learning hidden markov models.
\newblock In {\em Proceedings of the 22nd Conference on Learning Theory (COLT'09)}.

\bibitem[Julier and Uhlmann, 1997]{JU97}
Julier, S.~J. and Uhlmann, J.~K. (1997).
\newblock New extension of the {K}alman filter to nonlinear systems.
\newblock In {\em Proc.\@ Vol.\@ 3068, Signal Processing, Sensor Fusion, and Target Recognition VI}.

\bibitem[Julier and Uhlmann, 2004]{JU04}
Julier, S.~J. and Uhlmann, J.~K. (2004).
\newblock Unscented filtering and nonlinear estimation.
\newblock {\em Proc.\@ of the IEEE}, 92:401--422.

\bibitem[Kalman, 1960]{Kal60}
Kalman, R. (1960).
\newblock A new approach to linear filtering and prediction problems.
\newblock {\em Trans.\@ of the ASME, Journal of Basic Engineering}, 82:35--45.

\bibitem[Karl et~al., 2017]{KSBS17}
Karl, M., Soelch, M., Bayer, J., and van~der Smagt, P. (2017).
\newblock Deep variational {B}ayes filters: {U}nsupervised learning of state space models from raw data.
\newblock In {\em ICLR}.

\bibitem[Katayama, 2005]{Kat05}
Katayama, T. (2005).
\newblock {\em Subspace Methods for System Identification}.
\newblock Springer.

\bibitem[Katayama and Picci, 1999]{KP99}
Katayama, T. and Picci, G. (1999).
\newblock Realization of stochastic systems with exogenous inputs and subspace identification methods.
\newblock {\em Automatica}, 35(10):1635--1652.

\bibitem[Kawahara, 2016]{Kaw16}
Kawahara, Y. (2016).
\newblock Dynamic mode decomposition with reproducing kernels for {K}oopman spectral analysis.
\newblock In {\em Advances in Neural Information Processing Systems 29 (NIPS'16)}, pages 911--919.

\bibitem[Kawahara et~al., 2007]{KYM07}
Kawahara, Y., Yairi, T., and Machida, K. (2007).
\newblock A kernel subspace method by stochastic realization for learning nonlinear dynamical systems.
\newblock In {\em Advances in Neural Information Processing Systems 19}, volume~19, pages 665--672.

\bibitem[Kitagawa, 1993]{Kit93}
Kitagawa, G. (1993).
\newblock A monte carlo filtering and smoothing method for non-gaussian nonlinear state space models.
\newblock In {\em Proc.\@ of the 2nd U.S.-Japan Joint Seminar on Statistical Time Series}, pages 110--131.

\bibitem[Klus et~al., 2018]{KNK+18}
Klus, S., N\"{u}ske, F., Koltai, P., Wu, H., Kevrekidis, I., Sch\"{u}tte, C., and No\'{e}, F. (2018).
\newblock Data-driven model reduction and transfer operator approximation.
\newblock {\em Journal of Nonlinear Science}, 28(3):985--1010.

\bibitem[Klus et~al., 2020]{KSM20}
Klus, S., Schuster, I., and Muandet, K. (2020).
\newblock Eigendecompositions of transfer operators in reproducing kernel {H}ilbert spaces.
\newblock {\em Journal of Nonlinear Science}, 30:283--315.

\bibitem[Koopman, 1931]{Koo31}
Koopman, B. (1931).
\newblock Hamiltonian systems and transformation in {H}ilbert space.
\newblock {\em Proceedings of the National Academy of Sciences of the United States of America}, 17(5):315--318.

\bibitem[Kostic et~al., 2023]{KLNP23}
Kostic, V., Lounici, K., Novelli, P., and Pontil, M. (2023).
\newblock Sharp spectral rates for koopman operator learning.
\newblock In {\em Advances in Neural Information Processing Systems 36 (NIPS'23)}, pages 32328--32339.

\bibitem[Krishnan et~al., 2017]{KSS17}
Krishnan, R., Shalit, U., and Sontag, D. (2017).
\newblock Structured inference networks for nonlinear state space models.
\newblock In {\em Proc.\@ of the AAAI Conf.\@ on Artificial Intelligence (AAAI'17)}, pages 2101--2109.

\bibitem[Lindquist and Picci, 1996]{LP96}
Lindquist, A. and Picci, G. (1996).
\newblock Canonical correlation analysis, approximate covariance extension, and identification of stationary time series.
\newblock {\em Automatica}, 32(5):709--733.

\bibitem[Lindquistand and Picci, 1991]{LP91}
Lindquistand, A. and Picci, G. (1991).
\newblock A geometric approach to modelling and estimation of linear stochastic systems.
\newblock {\em Journal of Mathematical Systems, Estimation and Control}, 1(3):241--333.

\bibitem[Mauroy and Mezi\'{c}, 2018]{MM18}
Mauroy, A. and Mezi\'{c}, I. (2018).
\newblock Global computation of phase-amplitude reduction for limit-cycle dynamics.
\newblock {\em Chaos: An Interdisciplinary Journal of Nonlinear Science}, 28:073108.

\bibitem[Mezi\'{c}, 2005]{Mez05}
Mezi\'{c}, I. (2005).
\newblock Spectral properties of dynamical systems, model reduction and decompositions.
\newblock {\em Nonlinear Dynamics}, 41:309--325.

\bibitem[Mezi\'{c}, 2013]{Mez13}
Mezi\'{c}, I. (2013).
\newblock Analysis of fluid flows via spectral properties of the koopman operator.
\newblock {\em Annual Review of Fluid Mechanics}, 45:357--378.

\bibitem[Mollenhauer et~al., 2020]{MSKS20}
Mollenhauer, M., Schuster, I., Klus, S., and Sch\"{u}tte, C. (2020).
\newblock Singular value decomposition of operators on reproducing kernel {H}ilbert spaces.
\newblock In {\em Advances in Dynamics, Optimization and Computation}, pages 109--131. Springer International Publishing.

\bibitem[Rowley et~al., 2009]{RMB+09}
Rowley, C., Mezi\'{c}, I., Bagheri, S., Schlatter, P., and Henningson, D. (2009).
\newblock Spectral analysis of nonlinear flows.
\newblock {\em J.\@ of Fluid Mechanics}, 641:115--127.

\bibitem[Rozanov, 1967]{Roz67}
Rozanov, Y. (1967).
\newblock {\em Stationary Random Processes}.
\newblock Holden-Day.

\bibitem[Schmid, 2010]{Sch10}
Schmid, P. (2010).
\newblock Dynamic mode decomposition of numerical and experimental data.
\newblock {\em J.\@ Fluid Mechanics}, 656:5--28.

\bibitem[Shirasaka et~al., 2017]{SKN17}
Shirasaka, S., Kurebayashi, W., and Nakao, H. (2017).
\newblock Phase-amplitude reduction of transient dynamics far from attractors for limit-cycling systems.
\newblock {\em Chaos: An Interdisciplinary Journal of Nonlinear Science}, 27:023119.

\bibitem[Smola et~al., 2007]{SGSS07}
Smola, A., Gretton, A., Song, L., and Sch\"{o}lkopf, B. (2007).
\newblock A hilbert space embedding for distributions.
\newblock In {\em Proc. of the 17th Int'l Conf. on Algorithmic Learning Theory (ALT'07)}, pages 13--31.

\bibitem[Song et~al., 2010]{SBS+10}
Song, L., Boots, B., Siddiqi, S., Gordon, G., and Smola, A. (2010).
\newblock Hilbert space embeddings of hidden {M}arkov models.
\newblock In {\em Proceedings of the 27th International Conference on Machine Learning (ICML'10)}, pages 991--998.

\bibitem[Song et~al., 2009]{SHSF09}
Song, L., Huang, J., Smola, A., and Fukumizu, K. (2009).
\newblock Hilbert space embeddings of conditional distributions with applications to dynamical systems.
\newblock In {\em Proceedings of the 26th Annual International Conference on Machine Learning (ICML'09)}, pages 961--968.

\bibitem[Sriperumbudur et~al., 2008]{SGF+08}
Sriperumbudur, B., Gretton, A., Fukumizu, K., Lanckriet, G., and Sch\"{o}lkopf, B. (2008).
\newblock Injective {H}ilbert space embeddings of probability measures.
\newblock In {\em Proc.\@ of the 21st Ann.\@ Conf.\@ on Learning Theory (COLT'09)}, pages 111--122.

\bibitem[Takeishi et~al., 2017a]{TKY17b}
Takeishi, N., Kawahara, Y., and Yairi, T. (2017a).
\newblock Learning koopman invariant subspaces for dynamic mode decomposition.
\newblock In {\em Advances in Neural Information Processing Systems 30}, pages 1130--1140.

\bibitem[Takeishi et~al., 2017b]{TKY17a}
Takeishi, N., Kawahara, Y., and Yairi, T. (2017b).
\newblock Subspace dynamic mode decomposition for stochastic {K}oopman analysis.
\newblock {\em Physical Review E}, 96:033310.

\bibitem[Williams et~al., 2015]{WKR15}
Williams, M., Kevrekidis, I., and Rowley, C. (2015).
\newblock A data-driven approximation of the {K}oopman operator: {E}xtending dynamic mode decomposition.
\newblock {\em Journal of Nonlinear Science}, 25:1307--1346.

\bibitem[Wilson et~al., 2016]{WHSX16}
Wilson, A., Hu, Z., Salakhutdinov, R., and Xing, E. (2016).
\newblock Deep kernel learning.
\newblock In {\em Proc.\@ of the 19th Int'l Conf.\@ on Artificial Intelligence and Statistics (AISTATS'16)}, pages PMLR 51:370--378.

\bibitem[Wojtusch and von Stryk, 2015]{wojtusch2015humod}
Wojtusch, J. and von Stryk, O. (2015).
\newblock Hu{M}o{D}-a versatile and open database for the investigation, modeling and simulation of human motion dynamics on actuation level.
\newblock In {\em 2015 IEEE-RAS 15th international conference on humanoid robots (humanoids)}, pages 74--79. IEEE.

\end{thebibliography}

\begin{thebibliography}{99}
\bibitem[Bagheri, 2014]{bag2014}
Bagheri, S. (2014). Effects of weak noise on oscillating flows: Linking quality factor, floquet modes, and koopman spectrum. Physics of Fluids, 26(9):094104.

\bibitem[Becker et al., 2019]{BPG+19}
Becker, P., Pandya, H., Gebhardt, G., Zhao, C., Taylor, C., and Neumann, G. (2019). Recurrent kalman networks: Factorized inference in high-dimensional deep feature spaces. In Proc. of the 36th Int’l Conf. on Machine Learning, volume PMLR 97, pages 544–552.

\bibitem[Chui et al., 1982]{CWS82}
Chui, C., Ward, J., and Smith, P. (1982). Cholesky factorization of positive definite bi-infinite matrices. Numerical Functional Analysis and Optimization, 5(1):1–20.

\bibitem[Fukumizu et al., 2004]{Fukumizu2004DimensionalityRF}
Fukumizu, K., Bach, F. R., and Jordan, M. I. (2004). Dimensionality reduction for supervised learning with reproducing kernel hilbert spaces. J. Mach. Learn. Res., 5:73–99.

\bibitem[Fukumizu et al., 2013]{FSG13}
Fukumizu, K., Song, L., and Gretton, A. (2013). Kernel bayes’ rule: Bayesian inference with positive definite kernels. Journal of Machine Learning Research, 14(118):3753–3783.

\bibitem[Gebhardt et al., 2019]{GKN19}
Gebhardt, G., Kupcsik, A., and Neumann, G. (2019). The kernel kalman rule: Efficient nonparametric inference by recursive least-squares and subspace projections. Machine Learning, 108:2113–2157.

\bibitem[Hansen, 2006]{hansen2006cma}
Hansen, N. (2006). The CMA evolution strategy: A comparing review. Towards a new evolutionary computation: Advances in the estimation of distribution algorithms, pages 75–102.

\bibitem[Katayama, 2005]{Kat05}
Katayama, T. (2005). Subspace Methods for System Identification. Springer.

\bibitem[Takeishi et al., 2017]{TKY17a}
Takeishi, N., Kawahara, Y., and Yairi, T. (2017). Subspace dynamic mode decomposition for stochastic Koopman
analysis. Physical Review E, 96:033310.

\bibitem[Williams et al., 2015]{WKR15}
Williams, M., Kevrekidis, I., and Rowley, C. (2015). A data-driven approximation of the Koopman operator:
Extending dynamic mode decomposition. Journal of Nonlinear Science, 25:1307–1346.

\bibitem[Wojtusch et al, 2015]{wojtusch2015humod}
Wojtusch, J. and von Stryk, O. (2015). HuMoD-a versatile and open database for the investigation, modeling and simulation of human motion dynamics on actuation level. In 2015 IEEE-RAS 15th international conference on humanoid robots (humanoids), pages 74–79. IEEE.

\end{thebibliography}



\end{document}